\documentclass[10pt,twocolumn,letterpaper]{article}
\usepackage{iccv}              

\definecolor{iccvblue}{rgb}{0.21,0.49,0.74}
\usepackage[pagebackref,breaklinks,colorlinks,allcolors=iccvblue]{hyperref}
\newcommand\blfootnote[1]{%
  \begingroup
  \renewcommand\thefootnote{}\footnote{#1}%
  \addtocounter{footnote}{-1}%
  \endgroup
}

\title{A Constrained Optimization Approach for Gaussian Splatting from Coarsely-posed Images and Noisy Lidar Point Clouds}

\author{Jizong Peng$^{1*}$, \quad Tze Ho Elden Tse$^{2*}$, \quad Kai Xu$^2$, \quad Wenchao Gao$^1$, \quad Angela Yao$^{2}$\\
$^1$dConstruct Robotics \quad $^2$National University of Singapore\\
{\tt\small \texttt{\{jizong.peng,wehchao.gao\}@dconstruct.ai} \quad \texttt{\{eldentse,kxu,ayao\}@comp.nus.edu.sg}}
}

\usepackage{longtable}
\usepackage{booktabs}
\usepackage{amsmath,amsfonts,amssymb}
\usepackage[nice]{nicefrac}
\usepackage{xfrac}
\usepackage{todonotes}
\usepackage{breqn}
\usepackage{comment}
\usepackage{adjustbox}
\usepackage{multirow}
\usepackage{tikz}
\usepackage{pifont} 
\usepackage{bbding}
\usepackage{wasysym}
\usepackage{amssymb}
\usepackage[normalem]{ulem}
\usepackage[bottom]{footmisc}

\renewcommand{\rm}[1]{\textup{#1}}
\newcommand{\partialn}[2]{\nicefrac{\partial #1}{\partial #2}}

\newcommand{\real}[0]{\mathbb{R}}

\usepackage{soul} 
\usepackage{color} 

\newcommand{\jz}[1]{{\color{ForestGreen} #1}}

\begin{document}
\twocolumn[{%
\renewcommand\twocolumn[1][]{#1}%
\maketitle
\begin{center}
\vspace{-2em}
    \captionsetup{type=figure}
    $\vcenter{\hbox{\resizebox{\linewidth}{!}{\begin{tabular}[c]{@{}c@{}}
    \includegraphics[width=0.92\linewidth]{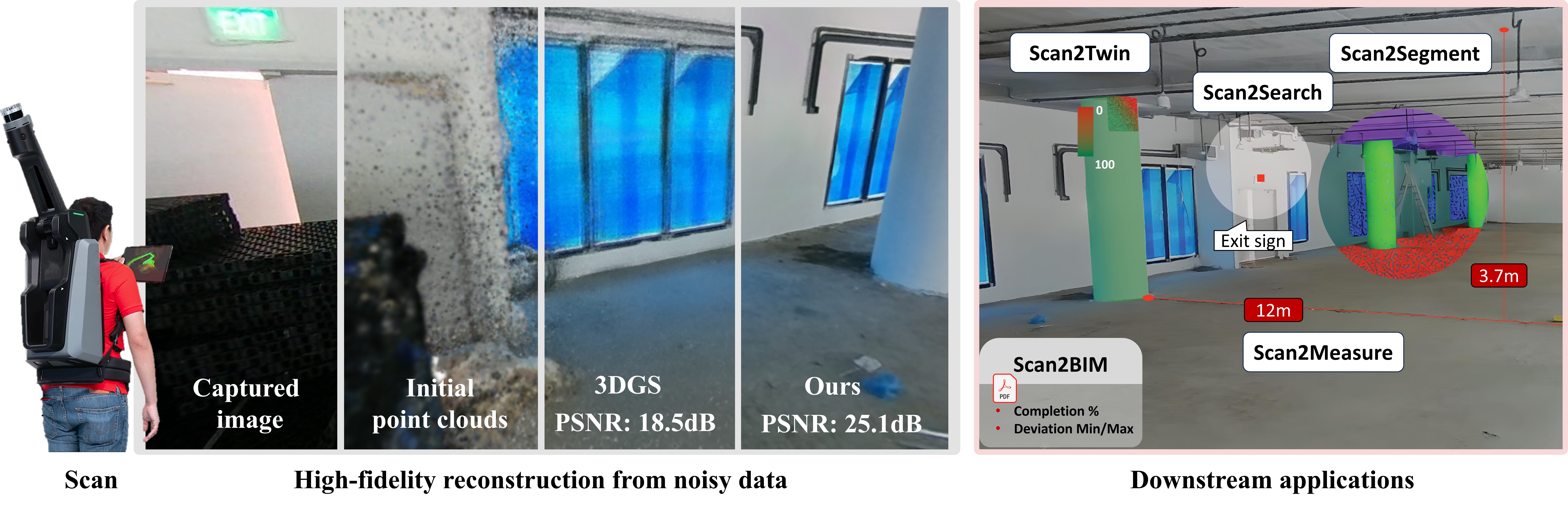} \\
    \end{tabular}}}}$
    \vspace{-0.7em}
    \captionof{figure}{
    Given noisy point clouds and inaccurate camera poses, our constrained optimization approach reconstructs the 3D scene in Gaussian Splatting with high visual quality, which enables various downstream applications. 
    }
    \label{fig:illustration}
\end{center}
}]
\maketitle
\begin{abstract}
3D Gaussian Splatting (3DGS) is a powerful reconstruction technique, but it needs to be initialized from accurate camera poses and high-fidelity point clouds. Typically, the initialization is taken from Structure-from-Motion (SfM) algorithms; however, SfM is time-consuming 
and restricts the application of 3DGS in real-world scenarios and large-scale scene reconstruction. 
We introduce a constrained optimization method for simultaneous camera pose estimation and 3D reconstruction that does not require S\textit{f}M support.  Core to our approach is decomposing a camera pose into a sequence of camera-to-(device-)center and (device-)center-to-world optimizations. 
To facilitate, 
we propose two optimization 
constraints conditioned to the sensitivity of each parameter group and restricts each parameter's search space. In addition, as we learn the scene geometry directly from the noisy point clouds, we propose geometric constraints to improve the reconstruction quality.
Experiments demonstrate that the proposed method significantly outperforms the existing (multi-modal) 3DGS baseline and methods supplemented by COLMAP on both our collected dataset and two public benchmarks.
\blfootnote{$^*$Equal contribution}
\vspace{-0.4cm}
\end{abstract}

\section{Introduction}
\label{sec:intro}
\begin{figure*}[t]
    \centering
    \includegraphics[width=0.95\linewidth]{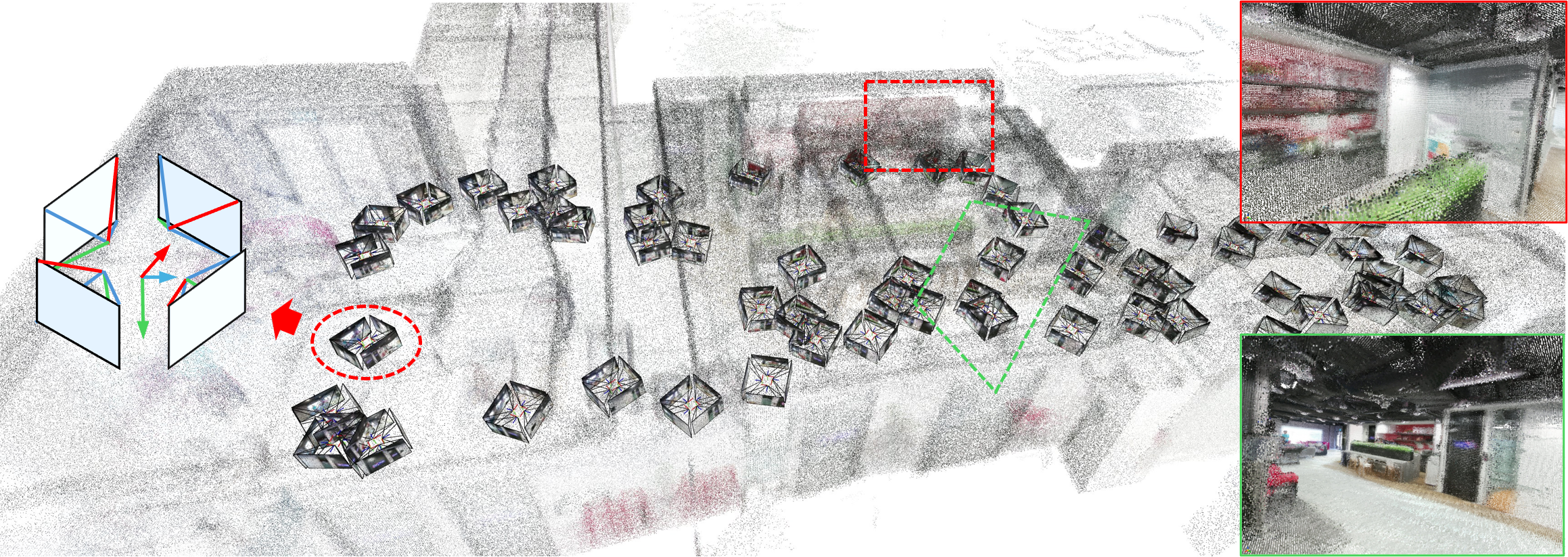}
    \caption{Qualitative example of camera poses and colored point clouds obtained from our multi-camera SLAM system. }
    \label{fig:context}
\end{figure*}

Simultaneous localization and mapping (SLAM) is critical for robotics and AR/VR applications. 
Traditional SLAM approaches~\cite{grisetti2010tutorial,mur2015orb,davison2007monoslam} are reasonably accurate in localization but struggle to produce dense 3D maps with fine-grained detailing. 
Recently, 3D Gaussian Splatting (3DGS)~\cite{kerbl20233d} has shown great promise for fast and high-quality rendering. As a result, there is increasing interest in combining 3DGS with SLAM~\cite{keetha2024splatam, deng2024compact, sun2024high, yan2024gs,matsuki2024gaussian}.  
One way is to incorporate SLAM for 3DGS initialization as a faster alternative to Structure-from-Motion (S\textit{f}M) algorithms.

Yet standard SLAM systems produce only rough camera pose estimates and noisy point clouds. 
Additionally, less-than-perfect camera intrinsics and Lidar-to-camera extrinsic calibration introduce errors and uncertainty into the 3D reconstruction. Directly using such SLAM inputs 
results in blurry reconstructions and degraded geometry (see Fig.~\ref{fig:illustration}) for standard 3DGS methods. 
While the SLAM outputs can be enhanced by additional hardware~\cite{jiang2024li,cui2024letsgo}, this invariably increases hardware costs and acquisition time. 


This paper addresses the challenge of training a 3DGS model under imprecise initialization conditions, including inaccurate sensor calibration and approximate camera pose estimation. We consider inputs from a typical 3D scanning setup, comprising multiple RGB cameras, a Lidar, and an inertial motion unit (IMU) within a rigid body framework. In the absence of S\textit{f}M support, we introduce a constrained optimization method for simultaneous camera estimation and 3D reconstruction. Specifically, our constrained optimization strategies are targeted to refine the extrinsics and intrinsics of the multi-camera setup, as well as the 3DGS.

To achieve this, we first decouple multi-camera poses into a sequence of camera-to-(device-) center and (device-) center-to-world transformations. However, simply optimizing for camera parameters and scene reconstruction can result in sub-optimal solutions for two main reasons.  First, there is inherent ambiguity in the perspective projection; the intrinsic parameters and camera poses describe relative and nonlinear relationships that can lead to multiple feasible solutions.  Secondly, the ensemble camera poses are over-parameterized; adjusting one camera’s orientation is equivalent to altering that of all device centers, creating unnecessary redundancy for optimization.

To address this problem, we pre-condition our optimization based on the sensitivity of each parameter group.  We also employ a log-barrier method to ensure that critical parameters remain within a predefined feasibility region (\eg focal length should not deviate by $2\%$). To further improve the quality of scene reconstructions, we propose two geometric constraints to serve as a strong regularization in the image space. Specifically, inspired by S\textit{f}M algorithms, we introduce a soft epipolar constraint and a reprojection regularizer 
for robust training to mitigate noisy camera poses.

There are no existing benchmarks fitting to this problem setting, so 
we curate a new dataset featuring complex indoor and large-scale outdoor scenes. As illustrated in Fig.~\ref{fig:context}, our proposed dataset are captured with 4 RGB cameras, an IMU and Lidar. 
We run an extensive ablation study as well as comparisons with state-of-the-art methods. Our experiments demonstrate that our constrained optimization approach is efficient and effective. 

In summary, our contributions are:
\begin{itemize}
    \item The first constrained optimization approach for training 3DGS that refines poor camera and point cloud initialization from a multi-camera SLAM system.
    \item We derive and enable refinement of camera intrinsics, extrinsics, and 3DGS scene representation using four of our proposed optimization constraints.
    \item A new dataset capturing complex indoor and large-scale outdoor scenes from hardware featuring multiple RGB cameras, IMU and Lidar.
    \item Our approach achieves competitive performance against existing 3DGS methods that rely on COLMAP, but with significantly less pre-processing time.
\end{itemize}


\section{Related Work}

\noindent \textbf{3D reconstruction.}
3D reconstruction from multi-view images is a fundamental problem in computer vision. Traditional methods use complex multi-stage pipelines involving feature matching, depth estimation~\cite{mi20223d}, point clouds fusion~\cite{chen2019point}, and surface reconstruction~\cite{kazhdan2013screened}.
In contrast, neural implicit methods like NeRF~\cite{mildenhall2021nerf} simplify this process by optimizing an implicit surface representation through volumetric rendering. 
Recent advancements include more expressive scene representations via advanced training strategies~\cite{chen2024nerf} and monocular priors~\cite{deng2022depth}.
However, these methods are often limited to foreground objects and are computationally intensive. More recently, 3DGS has been proposed as an efficient point-based representation for complex scenes.
While all the aforementioned methods require accurate camera poses, 3DGS also requires a geometrically accurate sparse point cloud for initialization.
This research addresses the challenges posed by inaccurate point clouds and camera poses to achieve a high-quality static reconstruction.

\noindent \textbf{Camera pose optimization.}
Recently, there has been growing interest in reducing the need for accurate camera estimation, often derived from S\textit{f}M. Initial efforts like i-NeRF~\cite{yen2021inerf} predicts camera poses by matching keypoints using a pre-trained NeRF. Subsequently, NeRF{\tiny$--$}~\cite{wang2021nerf} jointly optimizes the NeRF network and camera pose embeddings.  
BARF~\cite{lin2021barf} and GARF~\cite{chng2022gaussian} address the gradient inconsistency issue from high-frequency positional embeddings, with BARF using a coarse-to-fine positional encoding strategy for joint optimization.
In the 3DGS field, iComMa~\cite{sun2023icomma} employs an iterative refinement process for camera pose estimation by inverting 3DGS, while
GS-CPR~\cite{liu2024gsloc} uses visual foundation models for pose optimization with accurate key-point matches.
However, these methods assume a high-quality pre-trained 3DGS model and are computationally inefficient. In contrast, our method jointly optimize camera poses and reconstruction through constrained optimization.

\noindent \textbf{SLAM with 3DGS.}
The integration of 3DGS has gained significant interest in the field of SLAM~\cite{keetha2024splatam, deng2024compact, sun2024high, yan2024gs,matsuki2024gaussian}, serving as an efficient 3D scene representation.
Methods in this domain offer several advantages, such as continuous surface modeling, reduced memory usage, and improved gap filling and scene impainting for partially observed or occluded data.
In contrast, some work extend SLAM outputs to photometric reconstructions~\cite{cui2024letsgo,zhao2024tclc,zheng2024fast} by assuming accurate poses and point clouds due to complex hardware~\cite{zheng2024fast,cui2024letsgo} or multiple capture sequences~\cite{cui2024letsgo}. In this paper, we consider coarsely estimated poses and noisy point clouds from a multi-camera SLAM system to achieve highly accurate 3D scene reconstruction.



\noindent \textbf{Multimodal 3DGS.} 
There has been increasing interest in reconstruction with multimodal data~\cite{khan2024autosplat, lim2024lidar}, particularly for autonomous driving. For instance,~\cite{zhou2024drivinggaussian, yan2024street} combine images with Lidar, though they rely on COLMAP for refining camera poses. Additionally,~\cite{yan2024street} optimizes camera poses independently without intrinsic parameter refinement. In contrast, we are the first to introduce a constrained optimization framework that refines intrinsic and extrinsic parameters of (multiple) cameras under various constraints.

\section{Methodology}
In the following, we formulate our problem setting in Section~\ref{subsec:prelim} and detail how we enable intrinsic and extrinsic camera refinement in Section~\ref{subsec:refinement}. We then present our proposed optimization and geometric constraints in Section~\ref{subsec:optimization} Section~\ref{subsec:geometric}, respectively.

\subsection{Multi-camera problem setting} \label{subsec:prelim}
Given a set of coarsely estimated camera poses
\footnote{We refer to the camera pose as the \textit{camera-to-world} pose, indicating the camera’s position and orientation in world coordinates for simplicity.},
$\{\mathcal{P}_i\}|_{i=1}^{N} \in \mathbb{SE}(3)$, along with their respective RGB images $\{\mathcal{I}\}|_{i=1}^{N} \in \mathbb{R}^{H \times W \times 3}$, where $H$ and $W$ denote the height and width of the images, and $i$ represents the image/pose index ($1 \leq i \leq N$) among $N$ images. 
The poses are inaccurate due to two main reasons. Firstly, the orientation and position of the device $\hat{\mathcal{P}}_{i}$ derived from SLAM can be noisy due to sensor noise and drift in Lidar odometry estimation.
Secondly, the RGB images are captured 
asynchronously to the device pose acquisition. Specifically, the image pose $\mathcal{P}_i$ is roughly estimated by combining the closest device pose $\hat{\mathcal{P}}_i$ and the camera-to-device extrinsic $\mathcal{E}$.
This approach overlooks the inevitable time-frame offset (often up to 50 ms), further increasing the discrepancy between the estimated and true camera poses. In the following sections, we detail our approach to jointly correct the noisy set of camera poses and 3D point clouds within 3DGS scene representation. 

\subsection{Intrinsic and extrinsic refinement with 3DGS}\label{subsec:refinement}
\begin{figure}
    \centering
    \includegraphics[width=1\linewidth]{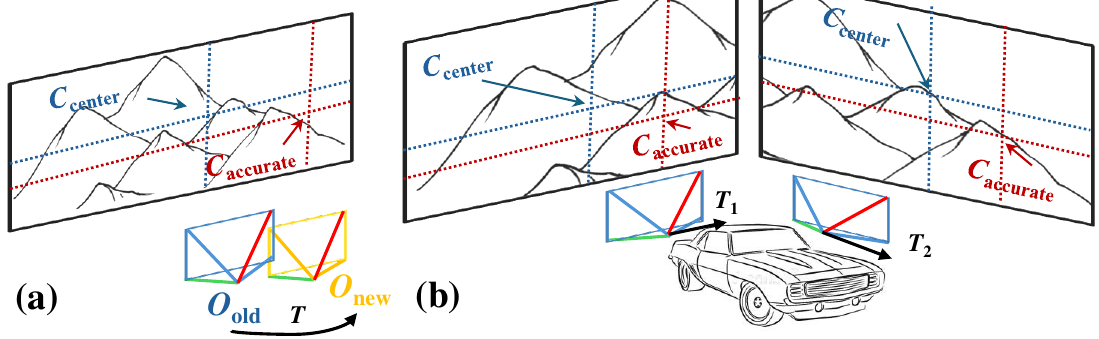}
    \caption{Illustration of camera intrinsic optimization.~(a) In monocular setting, inaccurate intrinsic parameters could be corrected by adjusting the camera pose,~\eg shifting the camera origin right by $T$.~(b) This approach is not feasible for multi-cameras under extrinsic constraints like autonomous cars or SLAM devices.
    }
    \label{fig:intrinsic_importance}
    \vspace{-0.4cm}
\end{figure}

\noindent \textbf{Intrinsic refinement via analytical solution.} Existing methods typically assume that camera intrinsics are provided~\cite{cui2024letsgo,zhao2024tclc} and 
overlook the importance of refining these parameters. 
As illustrated in Fig.~\ref{fig:intrinsic_importance}, the inaccuracies of camera intrinsics can be compensated via small extrinsic offsets for single-camera captures~\cite{yan2024gs,matsuki2024gaussian}. However, this approach fails in multi-camera systems (\eg SLAM or autonomous vehicles) where poses are constrained by the \underline{device} $\hat{\mathcal{P}}_{i}$. In multi-camera setups, inaccurate intrinsic parameters can significantly degrade rendered details, leading to blurry reconstructions. To enable intrinsic refinement, we apply the chain rule of derivation and obtain the analytical solutions for computing the gradient of each intrinsic parameters. We detail the derivation procedures in the Supplementary and provide qualitative examples of this enhancement in Fig.~\ref{fig:intrinsic}, showing improved image quality with clearer text.

\begin{figure}[!t]
    \centering
    \includegraphics[width=1.0\linewidth]{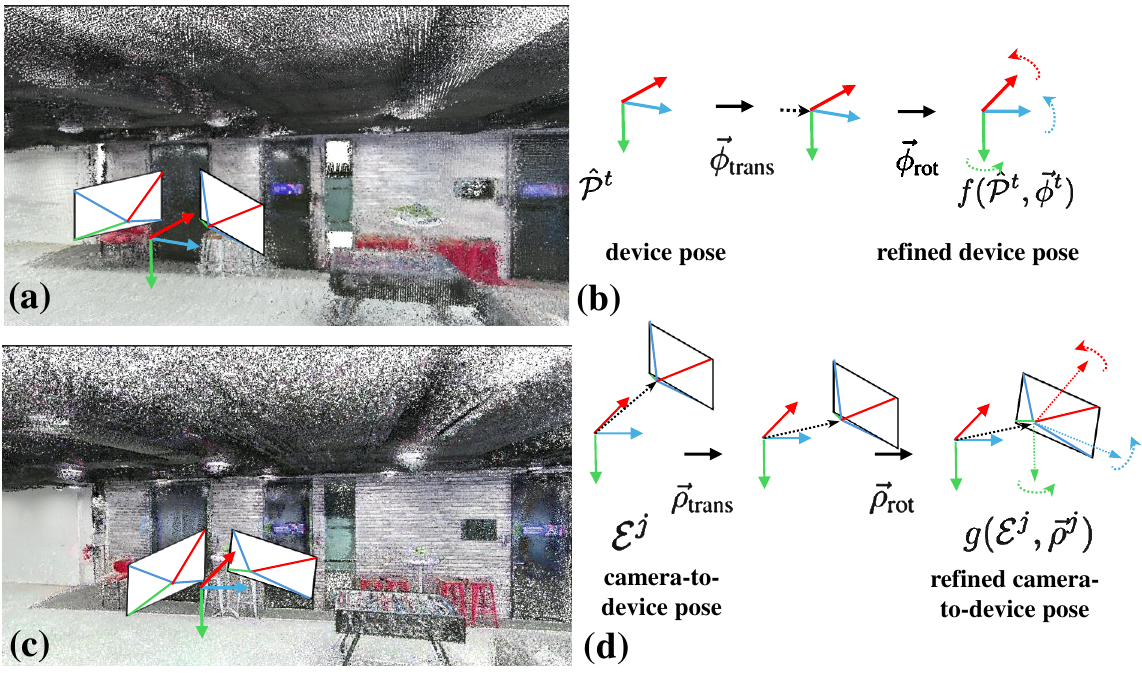}
    \caption{Illustration of our camera decomposition scheme. (a) Initial noisy point cloud from SLAM setup. (b) and (d) Optimization procedures of device-to-world and camera-to-device transformations. (c) Refined point cloud from our constrained optimization approach, showing improved visual quality.}
    \label{fig:camera-decomposition}
    \vspace{-0.4cm}
\end{figure}

\vspace{+1mm}
\noindent \textbf{Extrinsic refinement via camera decomposition.} 
Refining camera extrinsic in a multi-camera system is challenging due to the large number of parameters. For instance, a 4-camera rig with $10$k images involves $60$k degrees of freedom. To address this, we decompose each camera pose into two components: the camera-to-device pose and the device-to-world pose, expressed as: 
\begin{equation} \label{eq:decompose}
\vspace{-0.1cm}
    \mathcal{P}_{}^{(j,t)} = \hat{\mathcal{P}}^{t}_{} \times \mathcal{E}^{j}, 
    \vspace{-0.1cm}
\end{equation}
where $\mathcal{P}^{(j,t)}$ is the camera-to-world pose for camera $j$ at time $t$, $\hat{\mathcal{P}}^{t}$ is the device-to-world pose at $t$, and $\mathcal{E}^{j}$ is the camera-to-device extrinsic for camera $j$. This approach reduces the problem to modeling 4 shared extrinsics $\mathcal{E}^{j}$ and $2500$ independent device poses $\hat{\mathcal{P}}^{t}$, totaling $6\times 2500 + 6 \times 4 = 15024 $ degrees of freedom. Shared parameters across cameras and time frames simplify optimization and enhance the stability of joint camera pose refinement and accurate 3D scene reconstruction. This is illustrated in a real SLAM acquisition and its decomposition in Fig.~\ref{fig:camera-decomposition}.

We can now refine the camera extrinsics by applying small offsets to Eq.~\ref{eq:decompose}: 
\begin{equation}
{\mathcal{P}^{(j,t)}} = f({\hat{\mathcal{P}}^{t}}, \vec{\phi}^t)  \times g({\mathcal{E}}^{j}, \vec{\rho}^j),  
\vspace{-0.1cm}
\end{equation}
where $\vec{\phi}^t$ and $\vec{\rho}^j \in \mathbb{R}^6$ are learnable tensors, each consisting rotation $\vec{\phi}_{\rm{rot}}, \vec{\rho}_{\rm{rot}}\in \real^3 $ and a translation $\vec{\phi}_{\rm{trans}}, \vec{\rho}_{\rm{trans}}\in \real^3 $, to compensate for the device pose at time $t$ and the $j^{\rm{th}}$ camera-to-device error, respectively. Functions $f(\cdot)$ and $g(\cdot)$ define how these small deltas refine the noisy poses.

There are two general approaches to refine these poses. The first approach is to left-multiply the original pose by error matrix:
\begin{equation}
f({\hat{\mathcal{P}}^{t}}, \vec{\phi}^t)  = \underbrace{\Phi^{t}}_{\mathbb{SE}(3) \rm{ representation of } \phi^{t}}
\times \mathcal{\hat{P}}^t.
\label{equ:camera_pose_error_naive}
\vspace{-0.1cm}
\end{equation}
However, this leads to unstable optimization as it forces the camera location to rotate with respect to the world origin, which is often far from the initial camera value. To address this, we propose right-multiplying the error matrix with the original pose by defining the new device center as ${{\mathcal{P}}^{t}_{\rm{d2w}}}^{*}  =R_{\rm{d2w}}  \Delta t + t_{\rm{d2w}}$, and thus: 
\begin{equation}
f({\hat{\mathcal{P}}^{t}}, \vec{\phi}^t)  =  \mathcal{\hat{P}}^t 
\times \underbrace{\Phi^{t}}_{\mathbb{SE}(3) \rm{ representation of } \phi^{t}}.
\label{equ:camera_pose_error_propose}
\vspace{-0.1cm}
\end{equation}
We provide qualitative examples for these schemes in Supplementary and adopt the form in Eq.~\ref{equ:camera_pose_error_propose} for $f(\cdot)$ and $g(\cdot)$.

\subsection{Optimization constraints}\label{subsec:optimization}

Directly optimizing the camera parameters as formulated in Section~\ref{subsec:refinement} leads to sub-optimal solutions for two main reasons: 1) The inherent ambiguity in perspective projection, where intrinsic parameters and camera poses describe relative and nonlinear relationships, leading to multiple feasible solutions; and 2) The overparameterization of camera poses, where adjusting one camera’s orientation affects all device centers, creating unnecessary redundancy for optimization. In this section, we propose a \textit{sensitivity-based pre-conditioning} strategy to adjust the learning rate of each parameter and a \textit{log-barrier} strategy to constrain optimization within the feasible region.

\vspace{+1mm}
\noindent \textbf{Sensitivity-based pre-conditioning.} 
Inspired by the Levenberg-Marquardt algorithm, which is known for solving general nonlinear optimization problems like camera calibration~\cite{mittrapiyanuruk2006memo}, we propose an optimization approach that constrains parameter movements based on their \textit{sensitivity} and initial coarse estimates of poses and intrinsics. This has a strong motivation as a even a tiny refinement ($1\%$) in these parameters can lead to significantly different behaviors.

Given a dense point cloud $\mathcal{G}$, we render into $UV$ coordinates by camera-to-world $\mathcal{P}_{\rm{c2w}}$ and intrinsic $K$ matrices:
\begin{equation}\label{eq:proj}
\vspace{-0.1cm}
    (u, v) = {Proj}(  \phi_{\rm{rot}}, \phi_{\rm{trans}},   \rho_{\rm{rot}}, \rho_{\rm{trans}}
    |\mathcal{G}, \mathcal{P}_{\rm{c2w}}, K),
\end{equation}
where $Proj(\cdot)$ is the projection function.
We can then obtain the sensitivity matrix by solving the Jacobian of Eq.~\ref{eq:proj}:
\begin{dmath}
\mathcal{J}(\phi_{\rm{rot}}, \phi_{\rm{trans}}, \rho_{\rm{rot}}, \rho_{\rm{trans}} |\mathcal{G}, \mathcal{P}_{\rm{c2w}}, K
) 
= \\
\begin{pmatrix}
 \partialn{u}{\phi_{\rm{rot}}}& \partialn{u}{\phi_{\rm{trans}}} & \partialn{u}{\rho_{\rm{rot}}} & \partialn{u}{\rho_{\rm{trans}}}\\
 \partialn{v}{\phi_{\rm{rot}}}& \partialn{v}{\phi_{\rm{trans}}} & \partialn{v}{\rho_{\rm{rot}}} & \partialn{v}{\rho_{\rm{trans}}}
\end{pmatrix}.
\end{dmath}
The Jacobian matrix represents how small changes in each input components affect the output and can be efficiently computed. We take the average of individual $\mathcal{J}$ matrices for multi-view camera captures and adjust the learning rate based on the diagonal value ratio of $(\mathcal{J}^{\top}\mathcal{J})^{\nicefrac{-1}{2}}$, which is the inverse square root of the first-order approximation of the Hessian matrix. 

\vspace{+1mm}
\noindent \textbf{Log-barrier method to constrain the feasible region.} 
In addition to refining each parameter set with its sensitivity-based learning rate, we further construct a log-barrier constraint to ensure crucial parameters remain within their feasible boundaries by empirically assessing the error margin of each parameter.

\begin{figure}
    \centering
    \includegraphics[width=0.8\linewidth]{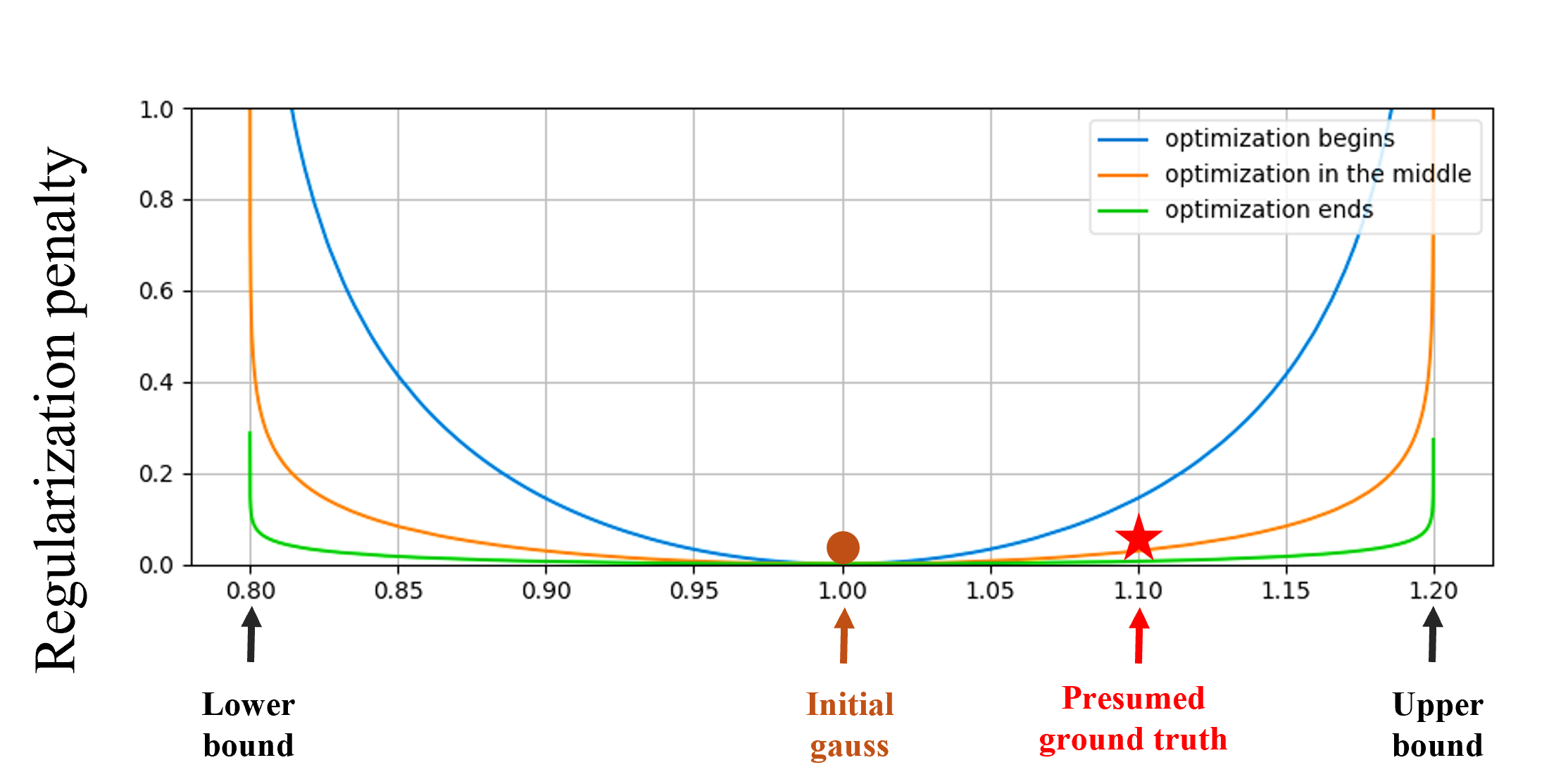}
    \caption{
    Illustration of the log-barrier method. Lower and upper bounds are predefined based on initial SLAM estimation. 
    At the start of the optimization, the barrier imposes a strong penalty for significant deviations from the initial estimate. As temperature increases, it transforms into a well-function, allowing the parameter to fully explore the feasible region.
    }
    \label{fig:log-barrier}
    \vspace{-0.4cm}
\end{figure}

To achieve this, we define $m$ inequality constraints $h_i(x)<0$, ($1\leq i \leq m$) for parameter $x$. The log-barrier method expresses these constraints in the negative log form, as $\mathcal{L}_{\rm{barrier}}=\nicefrac{1}{\mathcal{T}} \sum_{i=1}^{m} log (-h_i (x))$, where $\mathcal{T}$ is a temperature term that increases from a small value to a very large one. This formulation offers several advantages for training by inspecting the gradient of the negative log form:
\begin{equation}
    \frac{\partial \nicefrac{1}{\tau} log (-h_i(x))}{\partial x} = -\frac{1}{\tau h_i(x)} \frac{\partial h_i(x)}{x}.
    \vspace{-0.1cm}
\end{equation}
As shown in Fig.~\ref{fig:log-barrier}, 
this creates a symmetric penalty function centered around the initial value. The penalty gradient increases significantly as the parameter approaches the predefined boundaries because the gradient term $-\frac{1}{\tau h_i(x)}$ becomes large.  This prevents the parameter from entering infeasible regions. As optimization progresses, we increase the temperature $\mathcal{T}$ to reduce the penalty and allow the parameters to stabilize between the boundaries.
This design is ideal for our problem scenario as we can empirically set two bounds and guide the optimization toward a plausible solution. We apply these constraints to both the camera intrinsics and the decomposed camera pose transformations.

\subsection{Geometric constraints}\label{subsec:geometric}

In this section, we propose two geometric constraints to improve the robustness in mitigating noisy camera poses. We first use a state-of-the-art keypoint matching method~\cite{sun2021loftr}
to output semi-dense (up to several hundreds) keypoint matches $\{\vec{x}_{i}, \vec{x}_{i+n}\}$ for adjacent image frames $i$ and $i+n$. Here, $\vec{x}_{i}, \vec{x}_{i+n} \in \mathbb{R}^{M\times 2}$ represent $M$ matches for the image pair, and $n$ is a small integer $1\leq n\leq 3$ to ensure high co-visibility between images. The following two geometric constraints can effectively provide a strong prior for the relative poses between cameras in a multi-camera system.

\vspace{+1mm}
\noindent \textbf{Soft epipolar constraint.} 
This regularizes the learned relative camera poses to adhere the epipolar geometries. We implement this by first estimating the fundamental matrix $\mathbb{F}$, using the relative camera poses $\mathcal{P}_{i,j}$ and respective intrinsics $K_i$ and $K_j$, \ie $\mathbb{F}_{ij} = K_i ^{-\top} [t]_{\times} R_{ij}  K_j^{-1}.$

\noindent We can then compute the Sampson distance \cite{wang2023posediffusion} which takes the matched pixel pairs and $\mathbb{F}$ as inputs:
\begin{equation*}
\vspace{-0.1cm}
    \begin{gathered}
     \mathcal{L}_{\rm{epipolar}}(\vec{x}_i, \vec{x}_{i+n}, {\mathbb{F}}) = \\
\sum_{j=0}^{M-1} 
\frac{
    \vec{x}_{i+n}^{j^\top} \mathbb{F} \vec{x}_{i}^j
}{
    {\left( \mathbb{F} \vec{x}_{i}^j \right)}_1^2 + \left( \mathbb{F} \vec{x}_{i}^j \right)_2^2 + \left( \mathbb{F}^\top \vec{x}_{i+n}^j \right)_1^2 + \left( \mathbb{F}^\top \vec{x}_{i+n}^j \right)_2^2
}
.
    \end{gathered}
    \vspace{-0.1cm}
\end{equation*}

With this constraint as regularizer, we can achieve robust optimization convergence by incorporating prior information about camera intrinsics and extrinsics. However, since the epipolar constraint does not consider depth information and has projective ambiguities, we propose an additional geometric constraint in the following.

\vspace{+1mm}
\noindent \textbf{Reprojection error regularization.}
We extend the Bundle Adjustment from traditional S\textit{f}M algorithms into a geometric constraint that simultaneously optimizes both camera poses and 3DGS. This constraint can be expressed as:
\begin{dmath}
    \mathcal{L}_{\rm{reproj}}(\underbrace{\vec{x}_{i}, \vec{x}_{i+n}}_{\rm{matched points}}, \underbrace{\vec{d}_i, \vec{d}_{i+n}}_{\rm{depths}}| \underbrace{\mathcal{P}_{i}, \mathcal{P}_{i+n}}_{\rm{camera poses}}, \underbrace{K_i, K_{i+n}}_{\rm{intrinsics}}) \\
    = \sum_{j=0}^{M-1} (K_{i+n} \mathcal{P}_{i+n} \mathcal{P}_{i}^{-1} D_{i}^{j} K_i^{-1} \vec{x}_i^j - \vec{x}_{i+n}^{j}) +
\sum_{j=0}^{M-1} (K_{i} \mathcal{P}_{i} \mathcal{P}_{i+n}^{-1}  D_{i+n}^j K_{i+n}^{-1} \vec{x}_{i+n}^{j} - \vec{x}_i),
\label{equ:reprojection}
\end{dmath}
where $\vec{d}_i$ and $\vec{d}_{i+n} \in \real^{M\times 1}$ are the depths for the matched points in $i^{\rm{th}}$ and $i+n^{\rm{th}}$ images. This regularization term minimizes errors by considering depth distances, thus constraining the geometry of the scene which is complementary to the previous soft epipolar constraint.

Note that many existing works compute alpha-blending along the z-axis component of Gaussians in camera space to approximate \textit{rendered depth}. However, we found this approach unstable during optimization. Therefore, inspired by computer graphics, we instead compute line intersections to determine depths more accurately. We provide the mathematical derivation of this approach in the Supplementary. 

\section{Experiments}


\noindent \textbf{Implementation details.}
We train 3DGS using the following loss objective, which is a weighted combination of our proposed constraints and can be written as:
\begin{dmath}
    \mathcal{L}_{\rm{total}} = \underbrace{\mathcal{L}_{\rm{pixel}} + \lambda_{\rm{ssim}} \cdot \mathcal{L}_{\rm{ssmi}}}_{\rm{original learning objective}} + \underbrace{\lambda_{\rm{barrier}} \cdot \mathcal{L}_{\rm{barrier}}}_{\rm{log barrier constraint}}+ \underbrace{\lambda_{\rm{epi}}\cdot \mathcal{L}_{\rm{epipolar}} + \lambda_{\rm{reproj}}\cdot     \mathcal{L}_{\rm{reproj}}}_{\rm{geometry constraints}}.
    \label{eq:total}
\end{dmath}

We empirically set $\lambda_{\rm{ssim}} = 0.2$, $\lambda_{\rm{barrier}} = 0.1$, $\lambda_{\rm{epi}} = 1 \times 10^{-3}$ and $\lambda_{\rm{reproj}} = 5 \times 10^{-4}$ for Eq.~\ref{eq:total}. The smaller values for $\lambda_{\rm{epi}}$ and $\lambda_{\rm{reproj}}$ prevent significant deviations in relative poses due to noisy key-point matches.
We set the learning rate for intrinsic parameters to $8\times10^{-4}$. The base extrinsic learning rate is $5\times10^{-3}$, adjusted for each group of transformation parameters using the diagonal value ratios from $(\mathcal{J}^{\top}\mathcal{J})^{\nicefrac{-1}{2}}$. For log-barrier constraint on intrinsic parameters, we impose a strict bound of $\pm2\%$ deviation from the original value. We also apply adaptive constraints empirically for extrinsics: $\pm$0.625\textdegree and $\pm$2.5\textdegree for $\phi_{\rm{rot}}$ and $\rho_{\rm{rot}}$, and $\pm0.125m$ and $\pm0.5m$ for $\phi_{\rm{trans}}$ and $\rho_{\rm{trans}}$.
For all experiments, we follow~\cite{fan2024instantsplat} and adopt a \textbf{test-time adaptation strategy} on the unseen images to refine their camera poses. During test-time adjustments, we apply a learning rate of $5 \times 10^{-4}$ over 500 iterations while keeping the trained 3DGS parameters frozen. 
We apply this to the entire test set after training $48$k iterations. As most images are captured in uncontrolled settings with varying lighting and exposure~\cite{Reiser2023SIGGRAPH}, we introduce an efficient \textbf{exposure compensation module}.
We hypothesize that illumination variations are region-specific and affect image brightness gradually. Therefore, we correct this by a \textit{learnable low-frequency} offset. We detail this approach in the Supplementary.

\noindent \textbf{Dataset.} 
There is a lack of suitable public datasets of real-world multimodal SLAM sequences, which better reflect the challenges faced in industrial applications where scans are noisy and captured quickly. 
To address this, we collected data using our self-developed hardware across four scenes, including indoor and challenging outdoor settings. Our hardware, featuring four fisheye cameras, an IMU sensor, and a Lidar, scanned scenes such as a cafeteria, office room, laboratory (100-300$m^{2}$), and a residential district in East Asia (85$\times$45$m^{2}$). We present the key statistics of our dataset in Table~\ref{tab:dataset_stat}. Our captured dataset represents a unique problem setting and can be considered as a special case for autonomous driving. Specifically, as humans carry the capture device and walk around to capture the scene, it induces more significant vertical movements than typically autonomous driving datasets. In addition, these scans included significant lighting variations and moving people. Due to the absence of advanced hardware synchronization and sophisticated sensor calibration in our rapid data acquisition process, the resulting camera poses and point clouds from SLAM are particularly noisy around object surfaces. We provide details on our devices, acquisition protocol, and data pre-processing in the Supplementary and we will release this dataset upon paper acceptance.
We also benchmark on public datasets which are considered with less sensor noise: Waymo~\cite{sun2020scalability} for autonomous driving and GarageWorld~\cite{cui2024letsgo} for indoor measurement and inspection. 

\begin{table}[!h]
\centering
\caption{Key statistics of our proposed dataset. 
}
\begin{adjustbox}{max width=1\linewidth}
\setlength{\tabcolsep}{2pt}
\renewcommand{\arraystretch}{1}
\begin{tabular}{c|cccccc}
			\toprule
			\bfseries Scene &  \shortstack{scene\\type}
    &   \shortstack {frame \\ num} &  \shortstack{key-frame\\num}  &  \shortstack{scene\\dimension} 
    &  \shortstack{pcd\\size} \\
    \midrule
    Cafeteria & indoor & 5788 & 1260 &$20\times ~8$ & ~4 millions \\
    Office & indoor & 8184 & 1760 &$15\times 18$& 45 millions\\
    Laboratory & indoor & 5360 & 1216 &$15\times ~9$& 57 millions \\
    Town & outdoor & 8816 & 1932 &$85\times 45$& 39 millions\\
    
    \bottomrule
		\end{tabular}
	\end{adjustbox}
 \label{tab:dataset_stat}
\end{table}

%

\begin{table*}[!t]
\centering
\caption{Quantitative comparisons on our dataset. Red and blue highlights indicate the 1st and 2nd-best results, respectively, for each metric.~$^\triangle$ performs additional rig-based bundle adjustment to refine initial camera estimations. Our proposed method matches or surpasses the performance of the widely-adopt 3DGS-COLMAP approach while requiring significantly less data pre-processing time (prep. time).}
\label{tab:main_result}
\begin{adjustbox}{max width=1\textwidth}
\setlength{\tabcolsep}{3pt}
\renewcommand{\arraystretch}{1}
\begin{tabular}{l|c|ccc|ccc|ccc|ccc}
			\toprule
			\multirow[b]{2}{*}{\textbf{Methods}} 
   			& \multirow[b]{2}{*}{\textbf{Prep. time}} 
			& \multicolumn{3}{c|}{\textbf{Cafeteria}}
			& \multicolumn{3}{c|}{\textbf{Office}}
			& \multicolumn{3}{c|}{\textbf{Laboratory}}
			& \multicolumn{3}{c}{\textbf{Town}}\\
			\cmidrule(l{0pt}r{0pt}){3-5} 
			\cmidrule(l{0pt}r{0pt}){6-8} 
			\cmidrule(l{0pt}r{0pt}){9-11} 
			\cmidrule(l{0pt}r{0pt}){12-14} 
    & 
    & \textbf{PSNR} $\uparrow$ & \bfseries SSIM $\uparrow$ & \bfseries LPIPS $\downarrow$ 
    & \bfseries PSNR $\uparrow$ & \bfseries SSIM $\uparrow$ & \bfseries LPIPS $\downarrow$
    & \bfseries PSNR $\uparrow$ & \bfseries SSIM $\uparrow$ & \bfseries LPIPS $\downarrow$ 
    & \bfseries PSNR $\uparrow$ & \bfseries SSIM $\uparrow$ & \bfseries LPIPS $\downarrow$ 
     \\
    \midrule
    Direct reconst. & 3 minutes & 19.23 & 0.7887 & 0.2238 &17.49& 0.7577&0.2777&18.35 &0.7975 & 0.2207 &16.12 & 0.6151 & 0.3234\\
    Pose optimize. & 5 minutes & \cellcolor{blue!25}26.89 & \cellcolor{blue!25}0.8716 & \cellcolor{blue!25}0.1219 & 23.96 & 0.8366 & 0.1663 & 26.11 & 0.8673 & 0.1183 & 20.18 & 0.6845 & 0.2392 \\
    \midrule
			3DGS-COLMAP & 4-12 hours & 17.03 & 0.7681 & 0.2475 &\cellcolor{blue!25}25.82  & \cellcolor{blue!25} 0.8832 & \cellcolor{blue!25}0.1262& \cellcolor{blue!25} 28.30 &\cellcolor{blue!25} 0.9080 &  \cellcolor{red!25}0.0837
              &\cellcolor{blue!25}24.07 & \cellcolor{red!25} 0.8304 & \cellcolor{red!25} 0.1362      \\

   			3DGS-COLMAP$^\triangle$
			& 2-3 hours & 26.51 & 0.8379 & 0.1281 & 23.91& 0.8394 &0.1797 & 23.76 & 0.8157 & 0.1277 & 23.51 & 0.8090& 0.1534
			  \\

   \midrule
    CF-3DGS \cite{Fu_2024_CVPR} & 1 minute &  15.44 & 0.5412 & 0.5849 & 16.53 & 0.7555 & 0.4086 & 16.44 & 0.7557 & 0.3945 & 15.45 & 0.5412 & 0.5849\\
    MonoGS \cite{matsuki2024gaussian} & 1 minute & ~8.27 & 0.4684 & 0.6033  & ~9.56 & 0.4957 & 0.6560 & 13.08 & 0.6011 & 0.5103 & 12.74 & 0.3085 & 0.5331\\
    InstantSplat \cite{fan2024instantsplat} & $50$ minutes & 19.86 & 0.7743& 0.2548 & 23.30 & 0.8718 & 0.1451 & 20.89& 0.8624 & 0.1801 & 21.48 & 0.7378 & 0.2999\\
    
    \midrule
   
			Ours 
			& 5 minutes & \cellcolor{red!25}29.05 &\cellcolor{red!25}0.9168& \cellcolor{red!25}0.0817
   
			&\cellcolor{red!25} 26.07&\cellcolor{red!25}0.8850 & \cellcolor{red!25}0.1131 & 
    
    \cellcolor{red!25}28.64 & \cellcolor{red!25}0.9104 &\cellcolor{blue!25}0.0845 &
   \cellcolor{red!25}24.52 & \cellcolor{blue!25} 0.8259
			& \cellcolor{blue!25}0.1428 \\

			\bottomrule
		\end{tabular}
	\end{adjustbox}
\end{table*}

\begin{table*}
\centering
\caption{Quantitative comparisons on GarageWorld (\textit{left}) and Waymo (\textit{right}) datasets {with state-of-the-art multimodal methods}.}

\label{tab:public_result}

\begin{adjustbox}{max width=1\textwidth}
\setlength{\tabcolsep}{3pt}
\renewcommand{\arraystretch}{0.95}
\begin{tabular}{l|ccc|ccc|ccc|ccc|ccc}
			\toprule
                & \multicolumn{9}{c|}{\textbf{GarageWorld}~\cite{cui2024letsgo}} 
                & \multicolumn{6}{c}{\textbf{Waymo}~\cite{sun2020scalability}} \\ 
			\multirow[b]{2}{*}{\textbf{Methods}} 
			& \multicolumn{3}{c}{\textbf{Group 0}}
			& \multicolumn{3}{c}{\textbf{Group 3}}
			& \multicolumn{3}{c|}{\textbf{Group 6}}
			& \multicolumn{3}{c}{\textbf{Scene 002}}
			& \multicolumn{3}{c}{\textbf{Scene 031}}            \\
			\cmidrule(l{0pt}r{0pt}){2-4} 
			\cmidrule(l{0pt}r{0pt}){5-7} 
			\cmidrule(l{0pt}r{0pt}){8-10} 
			\cmidrule(l{0pt}r{0pt}){11-13} 
			\cmidrule(l{0pt}r{0pt}){14-16} 
    
    & \bfseries PSNR $\uparrow$ & \bfseries SSIM $\uparrow$ & \bfseries LPIPS $\downarrow$ 
    & \bfseries PSNR $\uparrow$ & \bfseries SSIM $\uparrow$ & \bfseries LPIPS $\downarrow$ 
    & \bfseries PSNR $\uparrow$ & \bfseries SSIM $\uparrow$ & \bfseries LPIPS $\downarrow$ 
    & \bfseries PSNR $\uparrow$ & \bfseries SSIM $\uparrow$ & \bfseries LPIPS $\downarrow$ 
    & \bfseries PSNR $\uparrow$ & \bfseries SSIM $\uparrow$ & \bfseries LPIPS $\downarrow$ 
     \\
    \midrule
    
			3DGS~\cite{kerbl20233d}  & \cellcolor{blue!25}25.43 & 0.8215 & \cellcolor{blue!25}0.2721 & 23.61  & 0.8162 & \cellcolor{blue!25}0.2698& 21.23 &0.7002  & 0.4640 &  25.84& 0.8700 & 0.1746 &  24.42 & 0.8328 &0.1783   \\

        LetsGo~\cite{cui2024letsgo} & 25.29 &\cellcolor{red!25}0.8387& 0.2978& \cellcolor{red!25}25.31& \cellcolor{red!25}0.8329& 0.2804 & \cellcolor{blue!25}21.72 & \cellcolor{blue!25}0.7462 & \cellcolor{blue!25}0.445 & 26.11 & 0.8429 & 0.2951 & 24.79 & 0.7851 & 0.3477\\

       Street-GS~\cite{yan2024street} & 24.20 & 0.8222& 0.2993& 24.19&0.8209&0.2849& 20.52& 0.7206& 0.4763&\cellcolor{blue!25} 27.96 &\cellcolor{blue!25} 0.8708 &\cellcolor{blue!25} 0.1664 & \cellcolor{blue!25}25.04 & \cellcolor{blue!25}0.8553 &\cellcolor{blue!25} 0.1697 \\

    \midrule
   
			Ours 
			& \cellcolor{red!25} 26.06 & \cellcolor{blue!25}0.8325 & \cellcolor{red!25}0.2605
   
			&\cellcolor{blue!25}25.07&\cellcolor{blue!25}0.8311 & \cellcolor{red!25} 0.2523& 
    
   \cellcolor{red!25} 23.76 &\cellcolor{red!25} 0.7779 & \cellcolor{red!25}0.3537 &  \cellcolor{red!25} 29.75 & \cellcolor{red!25}0.883& \cellcolor{red!25}0.161 & \cellcolor{red!25}28.48  & \cellcolor{red!25}0.868 & \cellcolor{red!25}0.1450 \\
			
			\bottomrule
		\end{tabular}
	\end{adjustbox}
\end{table*}

\noindent \textbf{Evaluation metrics.} 
Obtaining ground truth camera poses from real world settings is challenging. Consequently, existing works~\cite{mueller2022instant,Fu_2024_CVPR} often adopts COLMAP outputs as pseudo ground truth. However, as shown in Tab.~\ref{tab:main_result}, we demonstrate that COLMAP-generated poses are prone to failures, sometimes catastrophic, making it unreliable to use as ground truth for evaluation. This aligns with existing research showing that some approaches can be more accurate than COLMAP on individual scenes~\cite{brachmann2024acezero}, and evaluation rankings vary depending on the reference algorithm used for obtaining pseudo ground truths~\cite{brachmann2021limits}. 
We follow established methods~\cite{brachmann2024acezero,kerbl20233d,fan2024instantsplat} and assess pose quality in a self-supervised manner using novel view synthesis~\cite{waechter2017virtual}. 
Specifically, we sample test images at $N$ intervals, with $N$ determined per scene to ensure contains 60 testing images.
We report Peak Signal-to-Noise Ratio (PSNR), Structural Similarity Index Measure (SSIM) and Learned Perceptual Image Patch Similarity (LPIPS) to evaluate rendering quality.

\noindent \textbf{Comparison methods.} We compare our constrained optimization approach with various reconstruction methods, both with and without COLMAP, as well as SLAM-based Gaussian Splatting methods. We categorize them as follow:
\begin{itemize} [noitemsep,nolistsep]
    \item \textbf{Direct reconstruction}: This baseline directly optimizes scene reconstruction using the outputs from SLAM which include noise from various components. Therefore, this is considered the lower bound for our approach.
    \item \textbf{Pose optimization}: This baseline optimizes both the 3DGS parameters and the camera poses. It does not take into account of the multi-camera configuration and does not refine camera intrinsic parameters. This comparison method is commonly seen in incremental SLAM papers~\cite{lan2024monocular,matsuki2024gaussian,keetha2024splatam} and can be served as a strong baseline as it aligns with the learning objectives of the \textit{mapping} or \textit{global bundle adjustment} process.
    \item \textbf{3DGS-COLMAP}: The following two methods leverage COLMAP to derive precise camera poses. Despite time-consuming computation, COLMAP is widely adopted for training 3DGS as the resulting poses can often be considered as ground truth. We initially included this baseline as the performance upper bound. In the first variation, \textbf{3DGS-COLMAP} uses roughly estimated camera intrinsics to guide the optimization of camera poses. The subsequent variant, \textbf{3DGS-COLMAP$^{\triangle}$}, integrates additional approximate camera poses and refines them through a rig-based bundle adjustment (BA). This rig-based BA maintains a learnable, yet shared constant pose constraint across multiple cameras, making it the most relevant baseline for comparison.
    \item \textbf{Recent progress}: We compare with two SLAM-based 3DGS methods including CF-3DGS~\cite{Fu_2024_CVPR} and MonoGS~\cite{matsuki2024gaussian}. We also compare with InstantSplat~\cite{fan2024instantsplat} which uses a foundation model to provide relative poses and refine reconstruction geometry.
    \item \textbf{Multimodal 3DGS}: We compare with LetsGo~\cite{cui2024letsgo} and Street-GS~\cite{yan2024street} which takes Lidar data as input for large-scale public benchmarks. We provide implementation details of these methods in the Supplementary.

\end{itemize}

\begin{figure*}
\centering

    \includegraphics[width=1.0\linewidth]{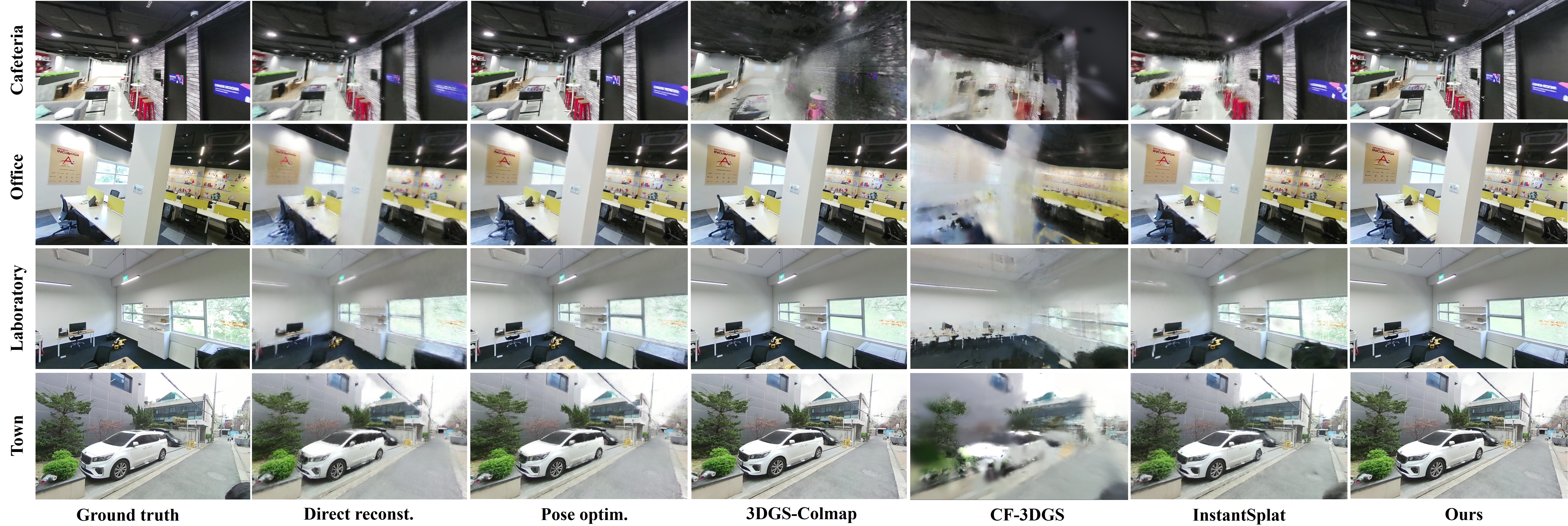}
    \caption{Qualitative comparisons with existing approaches. Our method achieves high rendering quality across diverse scenes.}
    \label{fig:enter-label}
\end{figure*}

\subsection{Experimental results - Tables~\ref{tab:main_result} and \ref{tab:public_result}}

\noindent \textbf{Direct baselines (Table~\ref{tab:main_result} rows 1-2).} We show that direct reconstruction using noisy SLAM outputs results in low rendering quality for all indoor/outdoor scenes. In contrast, pose optimization method shows slight improvements over the baseline with SSIM increases by 8.3\%, 7.89\%, 6.97\%, and 6.94\% for each of the scenes. Both methods underperformed in the Town scene due to its complex geometry and noisy point clouds.

\noindent \textbf{COLMAP-based methods (Table~\ref{tab:main_result} rows 3-4).} 3DGS-COLMAP is extensively applied to various 3D reconstruction tasks, yielding satisfactory results for three out of four datasets (SSIM: 0.88, 0.90, and 0.83) despite requiring up to 12 hours of computation time. However, it failed in the Cafeteria scene due to repetitive block patterns (details in Supplementary). In contrast, 3DGS-COLMAP$^\triangle$ has a reduced pose estimation time of 2-3 hours due to SLAM pose prior and Rig-BA. While it produces a more balanced rendering quality, it underperforms in the last two scenes compared to 3DGS-COLMAP, suggesting that rig optimization leads to sub-optimal outcomes.

\noindent \textbf{Recent progress (Table~\ref{tab:main_result} rows 5-7).} We show that both 3DGS for incremental SLAM methods, MonoGS and CF-3DGS, perform weakly across all evaluated datasets, with SSIM ranging from 0.40 to 0.75. This deficiency stems from their reliance on high quality image sequences, where accurate relative pose estimation depends heavily on image covisibility. Specifically, our dataset imposes a stringent 85\% covisibility threshold which makes it more challenging to obtain relative camera poses across the global scene. Additionally, the dataset contains various recurring block pattern as well as plain surfaces which can lead to degenerate solution. Conversely, InstantSplat achieves better rendering quality by leveraging foundation models.

\noindent \textbf{Multimodal 3DGS (Table~\ref{tab:public_result}).} Our approach achieve the best score in 12 cases and the second-best in the remaining ones. Notably, Street-GS also includes pose optimization, similar to our 3DGS-COLMAP baseline. However, our method shows significant improvement due to the combination of camera decomposition, intrinsic optimization, and various constraints, all without relying on COLMAP. We present additional quantitative analysis and qualitative comparisons on both public datasets in the Supplementary.

\subsection{Ablations}
\noindent \textbf{Camera decompositon \& pre-conditioning.} Directly optimizing camera parameters in a multi-camera setup can be computationally inefficient without improving reconstruction quality. To address this, we propose a camera decomposition and sensitivity-based pre-conditioning optimization strategies. As shown in Table~\ref{tab:cp-sp}, 
this approach achieves optimal performance with fast training convergence.


\noindent \textbf{Number of cameras.} 
\begin{table}
\centering
\caption{Ablations on number of cameras. We show that the improvement consistently increases with number of cameras.
}

\begin{adjustbox}{max width=0.99 \linewidth}
\setlength{\tabcolsep}{1.0pt}
\renewcommand{\arraystretch}{1}
\begin{tabular}{c|ccc|ccc|ccc}
			\toprule
			\multirow{2}{*}{\textbf{Methods}} 
			& \multicolumn{3}{c|}{\textbf{1 camera}}
			& \multicolumn{3}{c|}{\textbf{2 cameras}} 
			& \multicolumn{3}{c}{\textbf{4 cameras}} 
            \\
                \cmidrule(l{0pt}r{0pt}){2-10} 
			\bfseries     
    & \bfseries PSNR $\uparrow$ & \bfseries SSIM $\uparrow$   & \bfseries LPIPS $\downarrow$ 
    & \bfseries PSNR $\uparrow$ & \bfseries SSIM $\uparrow$  &\bfseries LPIPS $\downarrow$ & \bfseries PSNR $\uparrow$ & \bfseries SSIM $\uparrow$ &\bfseries LPIPS $\downarrow$ 
     \\
    \midrule 
     \multicolumn{10}{c}{\bfseries Cafeteria} \\
    \midrule
    Pose optim. & 27.51 & 0.881 & 0.079 &27.52 & 0.885&0.093 & 26.43 & 0.859 & 0.119  \\ 
    
    Ours  & 29.81 & 0.917  & 0.067  & 29.76 & 0.921 &0.072  &29.50 &0.922 & 0.077 \\
    \midrule
    Improv. & \,2.30 & 0.036 & 0.012  & 2.24 & 0.036 & 0.021 & \,\,3.07 & 0.063 & 0.042\\
\midrule

     \multicolumn{10}{c}{\bfseries Office} \\
     \midrule
    Pose optim. & 24.36 & 0.845 & 0.121 & 24.00& 0.832 & 0.141 & 23.38 & 0.827 & 0.169 \\ 
    
    Ours  &26.51 & 0.885&0.103  & 26.20 & 0.881 & 0.110& 26.12 & 0.891 & 0.109  \\
    \midrule
    Improv. & \,2.15 & 0.040 & 0.018  & 2.20 & 0.049 & 0.031 & \,\,2.74 & 0.064 & 0.060\\

    \bottomrule
		\end{tabular}
	\end{adjustbox}
 \label{tab:camera_num}
\end{table}
We evaluate the camera decomposition in Table~\ref{tab:camera_num} and show that our proposed method consistently improve the rendering quality. Our method is effective even in single-camera scenarios, as it links all camera poses with a shared camera-to-device matrix. This shared matrix provides a partial global constraint on the camera-to-device pose, simplifying the optimization process especially within limited training budgets. 

\begin{table}[!t]
\centering
\caption{Ablations on camera decomposition and sensitivity-based pre-conditioning strategies. \textbf{C.P.} and \textbf{P.C.} denote camera decomposition and pre-conditioning, respectively. In addition to standard rendering metrics, we report convergence percentage (\textbf{CVG\%}), indicating the training stage at which \textbf{SSIM} exceeds 95\% of its peak. A smaller values refers more stable optimization.
}
\begin{adjustbox}{max width=1\linewidth}
\setlength{\tabcolsep}{1pt}
\renewcommand{\arraystretch}{1}
\begin{tabular}{ll|cccc|cccc}
			\toprule
			\multicolumn{2}{c|}{\textbf{Methods}} 
			& \multicolumn{4}{c|}{\textbf{Cafeteria}}
			& \multicolumn{4}{c}{\textbf{Laboratory}}\\
                \cmidrule(l{0pt}r{0pt}){3-6} 
			\cmidrule(l{0pt}r{0pt}){7-10} 
			\bfseries C. D. & \bfseries P. C.    
    & \bfseries PSNR $\uparrow$ & \bfseries SSIM $\uparrow$ & \bfseries LPIPS $\downarrow$ & \bfseries CVG$\%$ 
    & \bfseries PSNR $\uparrow$ & \bfseries SSIM $\uparrow$ & \bfseries LPIPS $\downarrow$ & \bfseries CVG$\%$
     \\
    \midrule
    \ding{55} & \ding{55} & 26.91& 0.8659 & 0.1129 &34.38 & 27.00 & 0.8807 & 0.1045 & 31.25\\ 
    
    \ding{55} &\Checkmark  & 26.45 & 0.8577 & 0.1072 & \cellcolor{blue!25}22.92& 26.07 & 0.8645 & 0.1096 & \cellcolor{blue!25}18.76\\

    \Checkmark & \ding{55}& \cellcolor{blue!25}28.87& \cellcolor{blue!25}0.9154&\cellcolor{blue!25}0.0850 &43.10 &  \cellcolor{blue!25}28.52 & \cellcolor{blue!25}0.9092 & \cellcolor{blue!25}0.0894&  39.58\\
    \Checkmark & \Checkmark &\cellcolor{red!25}29.05 &\cellcolor{red!25}0.9168& \cellcolor{red!25}0.0817 & \cellcolor{red!25}15.65 & \cellcolor{red!25}28.64 & \cellcolor{red!25}0.9104 &\cellcolor{red!25}0.0845 &\cellcolor{red!25}16.67 \\
			\bottomrule
		\end{tabular}
	\end{adjustbox}
\label{tab:cp-sp}
   
\end{table}

\begin{figure}[!t]
    \centering
    \includegraphics[width=1\linewidth]{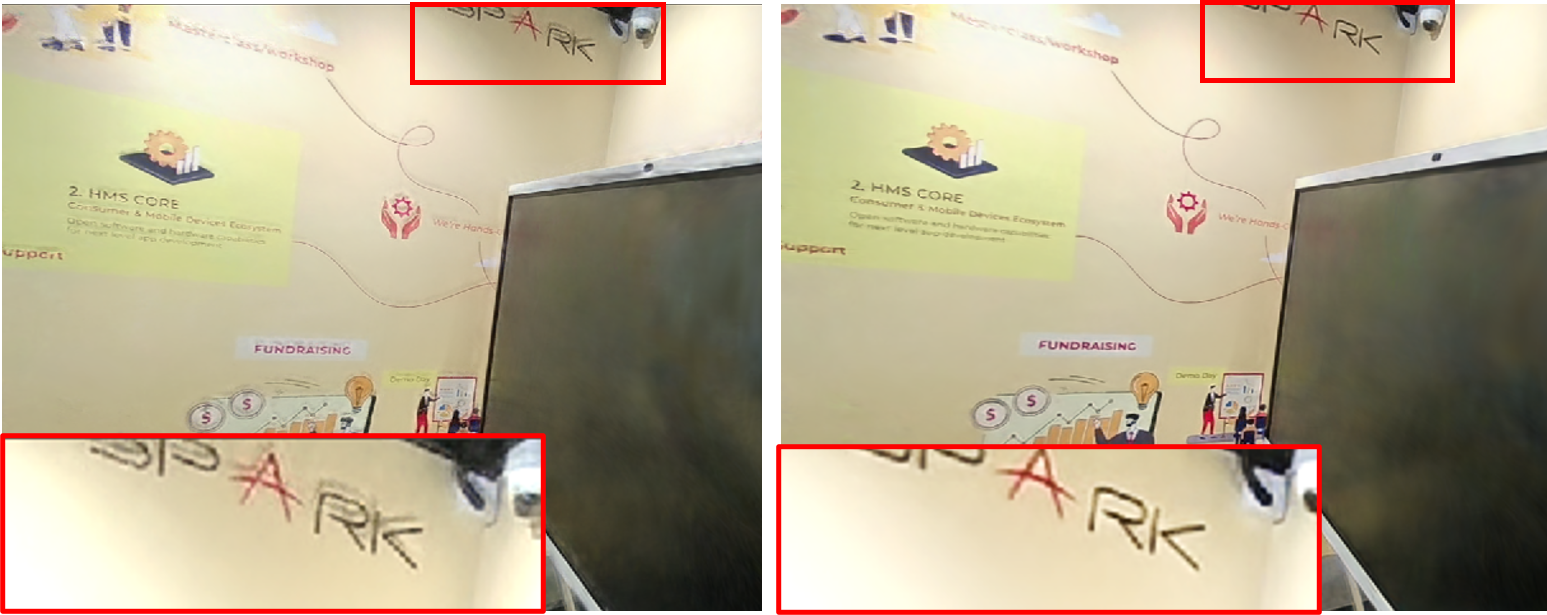}
    \caption{Qualitative examples for novel view synthesis with (\textit{right}) and without (\textit{left}) intrinsic refinement. We eliminate blurriness and enhance rendering quality by refining camera intrinsics during optimization.
    }
    \label{fig:intrinsic}
\end{figure}

\noindent \textbf{Intrinsic optimization.} Table~\ref{tab:intrinsic} shows that intrinsic refinement improve rendering quality, with consistent gains across all metrics.
In addition, we demonstrate that intrinsic refinement can deblur images by adjusting focal lengths and principal point in Fig.~\ref{fig:intrinsic}.

\begin{table}[t]
\centering
\caption{Ablations on intrinsic refinement.}
\begin{adjustbox}{max width=1\linewidth}
\setlength{\tabcolsep}{4pt}
\renewcommand{\arraystretch}{1}
\begin{tabular}{c|ccc|ccc}
			\toprule
			\multicolumn{1}{c|}{\textbf{Methods}} 
			& \multicolumn{3}{c|}{\textbf{Cafeteria}}
			& \multicolumn{3}{c}{\textbf{Laboratory}}\\
                \cmidrule(l{0pt}r{0pt}){1-4} 
			\cmidrule(l{0pt}r{0pt}){5-7} 
			\bfseries Refinement    
    & \bfseries PSNR $\uparrow$ & \bfseries SSIM $\uparrow$ & \bfseries LPIPS $\downarrow$  
    & \bfseries PSNR $\uparrow$ & \bfseries SSIM $\uparrow$ & \bfseries LPIPS $\downarrow$ 
     \\
    \midrule
    \ding{55} & 27.40 & 0.8975 & 0.0976 & 26.79 & 0.8843 & 0.0932\\ 
    
    \Checkmark  &\cellcolor{red!25}29.05 &\cellcolor{red!25}0.9168& \cellcolor{red!25}0.0817  & \cellcolor{red!25}28.64 & \cellcolor{red!25}0.9104 &\cellcolor{red!25}0.0845  \\
    \bottomrule
		\end{tabular}
	\end{adjustbox}
 \label{tab:intrinsic}
\end{table}

\begin{figure}[]
    \centering
    \includegraphics[width=0.49\textwidth]{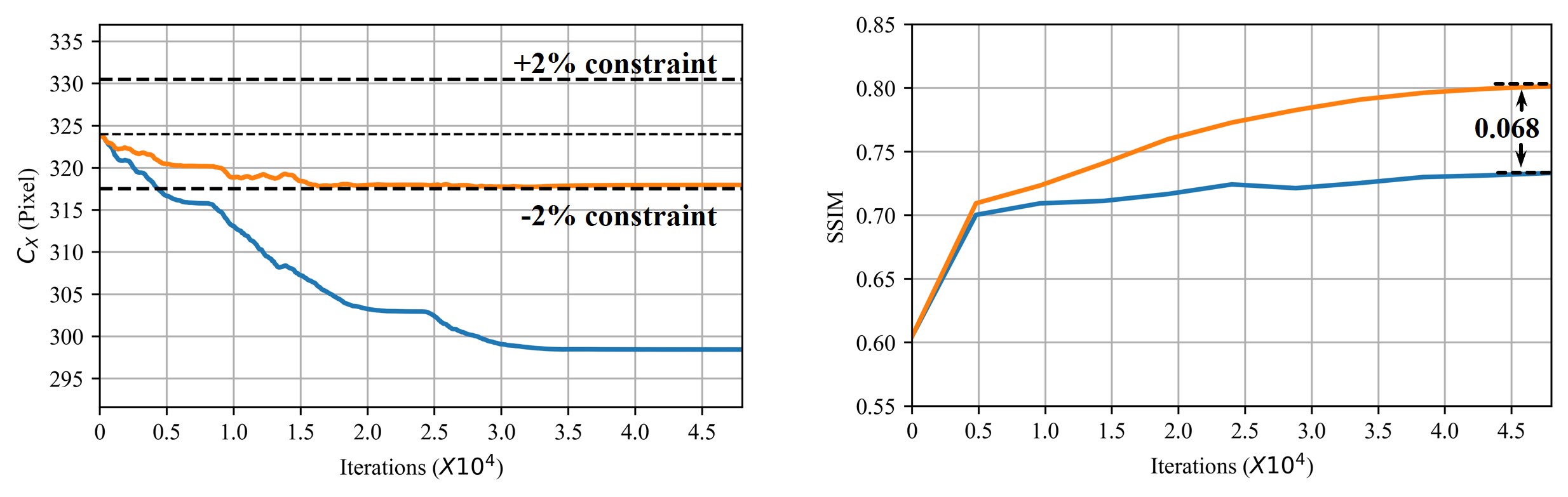}
    \caption{Ablations on log-barrier method. We show that training without log-barrier (blue plot) lead to significant principle point deviation (\textit{left}) and sub-optimal solution (\textit{right}). In contrast, using log-barrier (orange plot) results in higher SSIM (right).
    }
    \label{fig:log-barrier-result}
\end{figure}

\noindent\textbf{Log-barrier method.} Using only pre-conditioning optimization strategy is insufficient to prevent sensitive parameters to exceed their feasible region. To address this, we use a log-barrier method to constrain the feasible region. We show that by simply constraining the feasible region within $\pm 2\%$ improves SSIM by 6.8\% in Fig.~\ref{fig:log-barrier-result}.


\noindent\textbf{Geometric constraints.} We next asses the importance of the two proposed geometric constraints. In addition to standard metrics, we report the mean epipolar line error (Ep-e) and the reprojection error (RP-e) in Table~\ref{tab:geometry}.
We observe consistent performance gains with both geometric constraints, even as random noise increases in both device and camera-to-device poses. In addition, we provide qualitative examples on key-point matches and their corresponding epipolar lines in Fig.~\ref{fig:epipolar}. We show that minor epipole displacements due to geometric constraints significantly reduce epipolar line error from 2.70 to 0.75 pixels.

\begin{table}[!t]
\centering
\caption{Ablation study on geometric constraint. Ep-e stands for 
 mean epipolar line error (Ep-e) and RP-e denotes mean re-projection error. Our proposed losses helps to reduce both errors and increase the rendering quality.}
\label{tab:geometry}
\begin{adjustbox}{max width=1\linewidth}
\setlength{\tabcolsep}{4pt}
\renewcommand{\arraystretch}{1}
\begin{tabular}{c|cc|ccccc}
			\toprule
    \multirow{2}{*}{\begin{minipage}{0.3in} \textbf{Noise Level} \end{minipage}}
    & \multicolumn{2}{c|}{\textbf{Methods}} 
			& \multicolumn{5}{c}{\textbf{Cafeteria}}
			\\
    \cmidrule(l{0pt}r{0pt}){2-8} 
    & \bfseries E.P. & \bfseries R.P.    
    & \bfseries PSNR $\uparrow$ & \bfseries SSIM $\uparrow$ & \bfseries LPIPS $\downarrow$
    & \bfseries Ep-e $\downarrow$ &\bfseries RP-e $\downarrow$\\
    \midrule
    \multirow{4}{*}{\textbf{-}} & \ding{55} & \ding{55}  & 27.05 & 0.8945 & 0.1047 &1.14 &2.52\\ 
    
    &\ding{55} &\Checkmark & 27.24 & 0.9130 & 0.0906 & 1.11 & \cellcolor{blue!25}2.04 \\

    &\Checkmark & \ding{55}& \cellcolor{blue!25}27.25 & \cellcolor{blue!25}0.9141 & \cellcolor{blue!25}0.0895 & \cellcolor{blue!25}1.09 & 2.05\\
    & \Checkmark & \Checkmark &  \cellcolor{red!25}27.31 & \cellcolor{red!25}0.9147 &\cellcolor{red!25}0.0891 & \cellcolor{red!25}1.08 & \cellcolor{red!25}1.88\\

    \midrule

    \multirow{4}{*}{\textbf{0.2\textdegree}} & \ding{55} & \ding{55} & 26.04 & 0.8901 & 0.1007 & 1.23 & 2.56\\ 
    
    &\ding{55} &\Checkmark  & 26.16 & 0.8952 & 0.0989 & 1.17 &2.19\\

    &\Checkmark & \ding{55}& \cellcolor{blue!25}26.51 & \cellcolor{blue!25}0.9007 & \cellcolor{blue!25}0.0963 & \cellcolor{blue!25}1.12 & \cellcolor{blue!25}2.06\\
    & \Checkmark & \Checkmark & \cellcolor{red!25}26.84 & \cellcolor{red!25} 0.9045 & \cellcolor{red!25}0.0958 & \cellcolor{red!25}1.11 & \cellcolor{red!25}2.00\\

    \midrule

    \multirow{4}{*}{\textbf{0.5\textdegree}} & \ding{55} & \ding{55} & 24.80 & 0.8584 & 0.1244 & 1.72 & 3.92\\ 
    
    &\ding{55} &\Checkmark  & 24.87 & 0.8607 & 0.1196 & 1.42 & 2.99\\

    &\Checkmark & \ding{55}& \cellcolor{blue!25}25.18 & \cellcolor{blue!25}0.8665 & \cellcolor{blue!25}0.1138 & \cellcolor{blue!25}1.23 & \cellcolor{blue!25}2.35\\
    & \Checkmark & \Checkmark & \cellcolor{red!25}25.20 & \cellcolor{red!25} 0.8672 & \cellcolor{red!25}0.1120 & \cellcolor{red!25}1.21 & \cellcolor{red!25}2.32\\
			\bottomrule
		\end{tabular}
	\end{adjustbox}
\end{table}

\begin{figure}
\centering
\begin{adjustbox}{max width=1\linewidth}
    \centering
    \setlength{\tabcolsep}{1.1pt}
    \begin{tabular}{lcc}
        & Paired image 1 & Paired image 2\\
        
       \rotatebox{90}{~~ Coarse Pose}  & \includegraphics[width=0.5\linewidth]{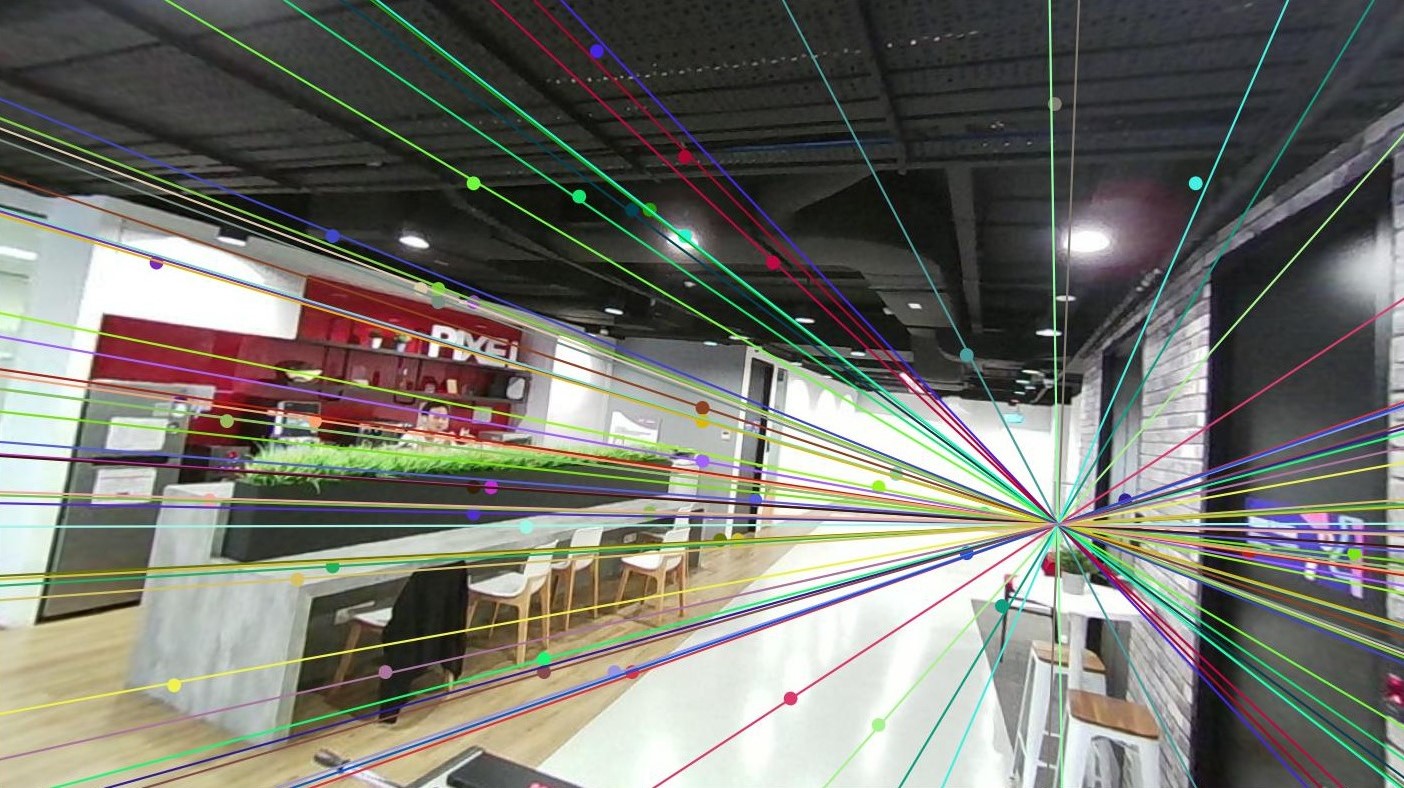} &\includegraphics[width=0.5\linewidth]{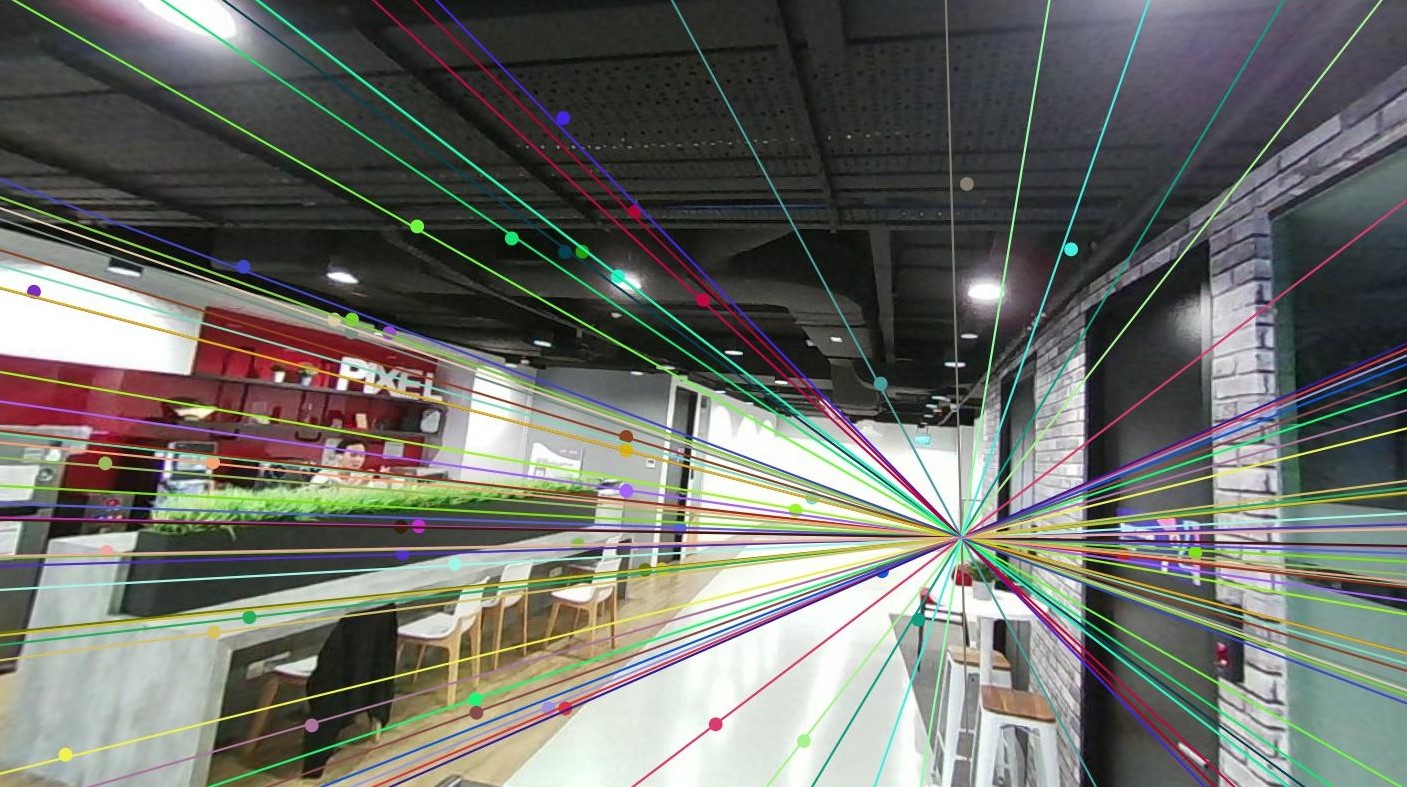}\\
       \midrule
       \rotatebox{90}{~~Refined Pose} & \includegraphics[width=0.5\linewidth]{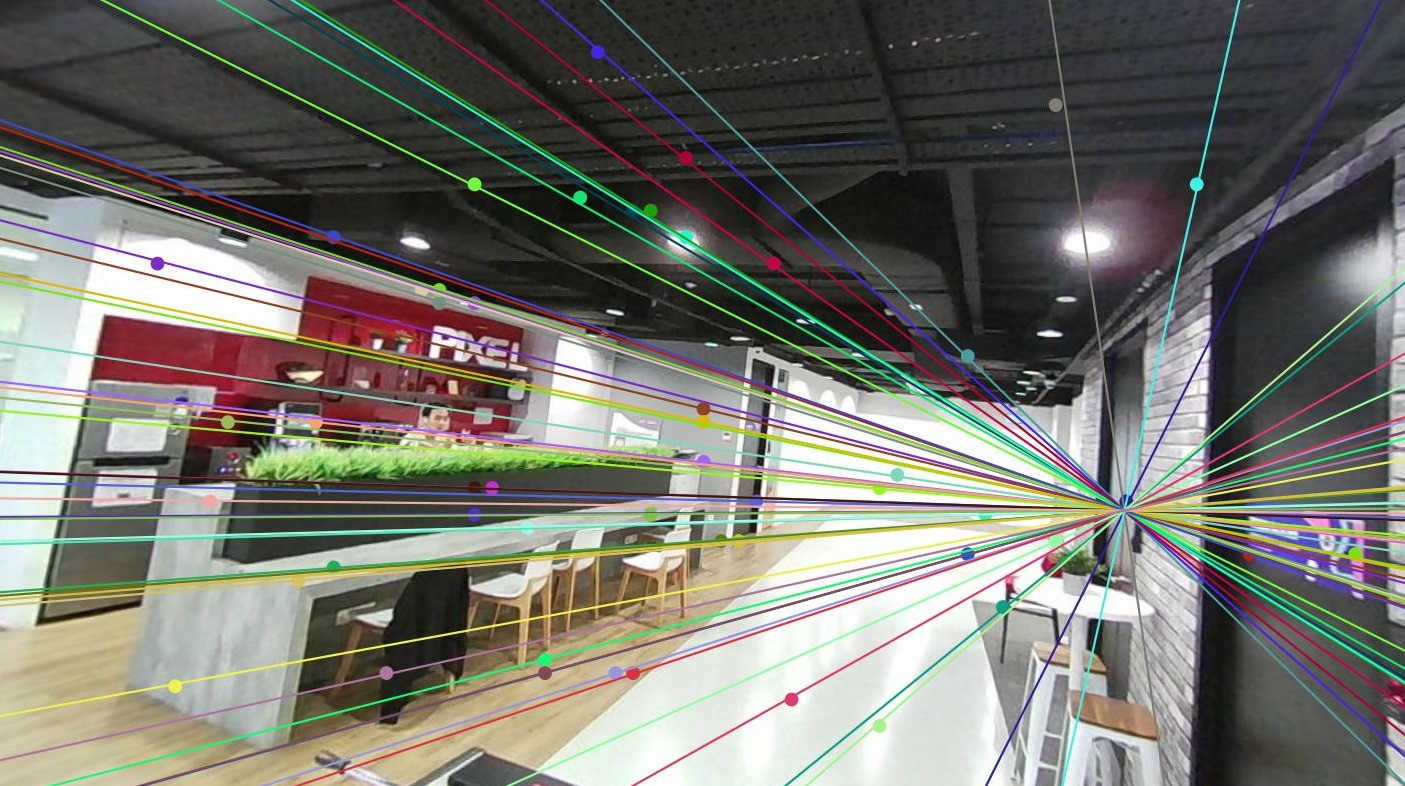} &\includegraphics[width=0.5\linewidth]{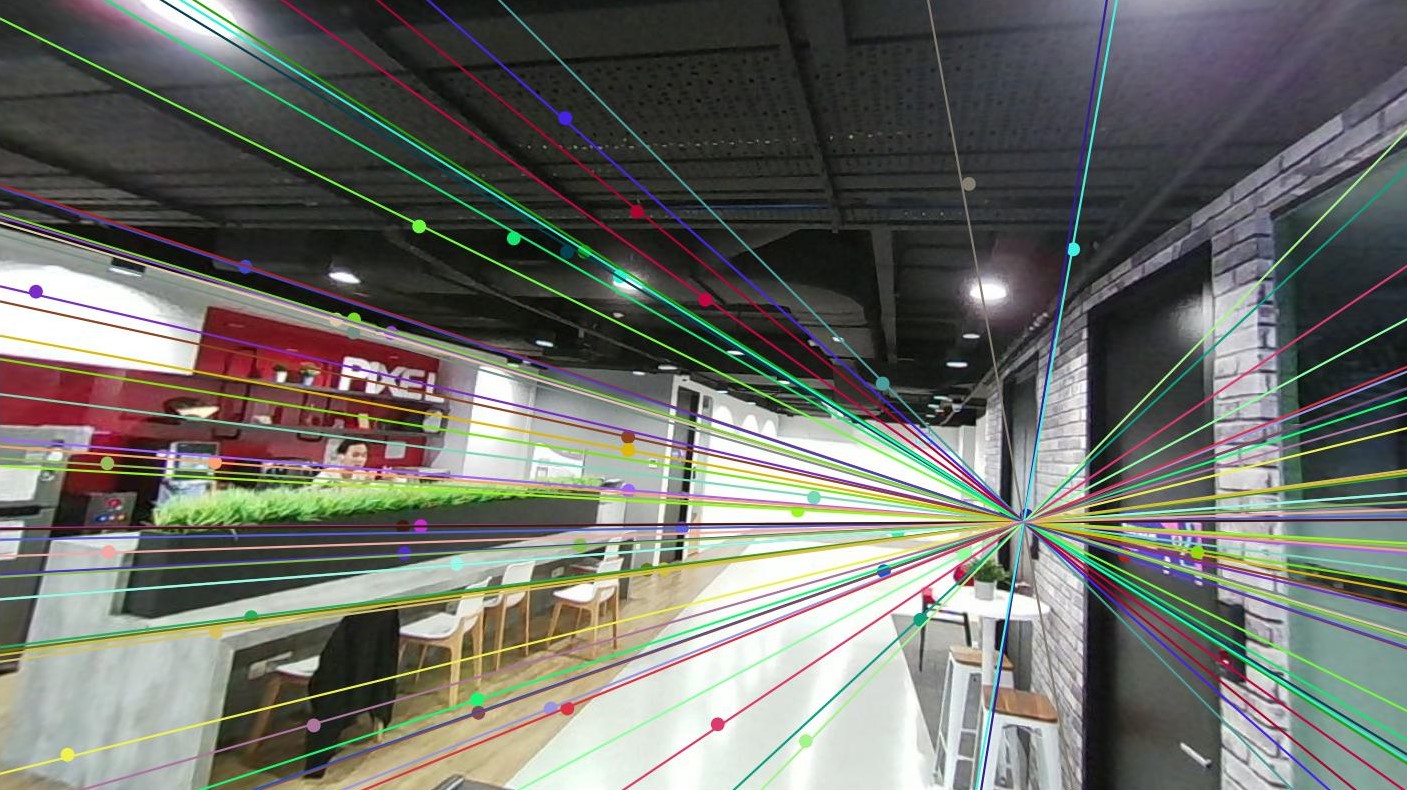}\\
       
    \end{tabular}
\end{adjustbox}
    
\caption{Qualitative examples on key-point matches and their corresponding epipolar lines. Vertical inspection shows that the geometric constraints cause minor epipole displacements towards lower epipolar error as well as better reconstruction quality.
}
\label{fig:epipolar}
\end{figure}

\noindent\textbf{Ablations on test-time adaptation}
For each test image, we keep the 3DGS parameters constant while refining the camera pose and learning the exposure compensation with low-frequency offset. Table~\ref{tab:apd-tta} provides a detailed analysis of the impact these components have on the experimental results.
Without applying either technique, we observe poor quality in novel-view renderings, primarily due to camera pose mismatches. 
Test-time pose optimization helps align the rendered image with the actual ground truth image, improving all visual metrics across both evaluated scenes, particularly the Cafeteria scene.
On the other hand, using only exposure compensation did not significantly enhance metrics related to visual and structural similarities (SSIM and LPIPS), though it did moderately increase the Peak Signal-to-Noise Ratio (PSNR). As expected, this exposure correction module addresses exposure errors but struggles to capture high-frequency details. Combining both techniques results in the best reconstruction performance in the tested scenes.

\begin{table}[!h]
\centering
\caption{Ablations on test-time adaptation. \textbf{Pose} denotes pose refinement while \textbf{Expo.} represents exposure correction module.}
\begin{adjustbox}{max width=1\linewidth}
\setlength{\tabcolsep}{2pt}
\renewcommand{\arraystretch}{1}
\begin{tabular}{cc|ccc|ccc}
			\toprule
			\multicolumn{2}{c|}{\textbf{Methods}} 
			& \multicolumn{3}{c|}{\textbf{Cafeteria}}
			& \multicolumn{3}{c}{\textbf{Laboratory}}\\
                \cmidrule(l{0pt}r{0pt}){1-5} 
			\cmidrule(l{0pt}r{0pt}){6-8} 
			\bfseries Pose & \bfseries Expo.    
    & \bfseries PSNR $\uparrow$ & \bfseries SSIM $\uparrow$ & \bfseries LPIPS $\downarrow $ 
    & \bfseries PSNR $\uparrow$ & \bfseries SSIM $\uparrow$ & \bfseries LPIPS $\downarrow$ \\
    \midrule
    \ding{55} & \ding{55}&  19.80 & 0.7752 & 0.1144 & 22.52 & 0.8984 &0.0939\\ 
    \ding{55} & \Checkmark & 22.65 & 0.7872& 0.1102 & \cellcolor{blue!25} 27.93 & \cellcolor{blue!25} 0.9065 & \cellcolor{blue!25}0.0881\\ 
    \Checkmark &\ding{55}  & \cellcolor{blue!25}23.04 & \cellcolor{blue!25}0.9026 & \cellcolor{blue!25}0.0876 & 22.67 & 0.9017 & 0.0933\\
    \Checkmark & \Checkmark  & \cellcolor{red!25} 28.58 & \cellcolor{red!25} 0.9156 & \cellcolor{red!25} 0.0820 & \cellcolor{red!25}28.18 & \cellcolor{red!25}0.9101 & \cellcolor{red!25}0.0875\\
    \bottomrule
		\end{tabular}
	\end{adjustbox}
 \label{tab:apd-tta}
 \end{table}
\section{Conclusion}
This paper presented a method for 3D Gaussian Splatting with noisy camera and point cloud initializations from a multi-camera SLAM system.  We proposed a
constrained optimization framework that decomposes the camera pose into camera-to-device and device-to-world transformations.  By optimizing each of these transformations individually under soft constraints, we can efficiently and accurately construct 3DGS representations.  We also introduced a new multi-view 3D dataset captured under these noisy albeit practical settings, which we will release to the community to encourage further development in this area of research.


\clearpage
{
    \small
    \bibliographystyle{ieeenat_fullname}
    \bibliography{main}
}
\clearpage
\section{Appendix}
\appendix

\noindent In this supplemental document, we provide:
\begin{itemize}
    \item details of dataset acquisition and pre-processing (Sec~\ref{apd:dc-dataset});
    \item derivation of intrinsic refinement (Sec~\ref{apd:intrinsic});
    \item details of exposure compensation module (Sec~\ref{apd:exposure});
    \item details of line intersection-based depth estimation (Sec~\ref{subsec:line_intersec});
    \item extended implementation details and discussion (Sec~\ref{apd:extended});
    \item supplementary experimental results on GarageWorld dataset (Sec~\ref{sec:letsgo});
    \item additional qualitative comparison on Waymo dataset (Sec~\ref{sec:waymo}).
\end{itemize}

\section{Dataset acquisition and pre-processing}\label{apd:dc-dataset}

\begin{figure}[b]
    \centering
    \includegraphics[width=1\linewidth]{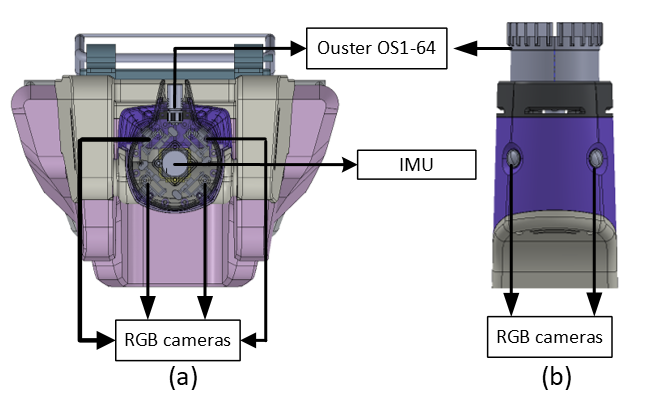}
    \caption{Key components of our SLAM hardware setup. The device includes a 64-channel mechanical Lidar on top, with four RGB cameras positioned at various angles to provide a complete 360-degree view. An IMU sensor is located directly beneath the Lidar. (a) System top-down view (b) front view.}
    \label{sensor_setup}
\end{figure}

In this section, we provide the configuration, calibration, and synchronization details of our SLAM hardware setup
as well as the main pre-processing steps for the captured images.

\paragraph{Data acquisition:}\label{subsec:data_acquisition}
As illustrated in Fig.~\ref{sensor_setup}, Our self-developed device comprises an Ouster OS1-64 Lidar, four Decxin AR0234 cameras with wide-angle lenses, and a Pololu UM7 IMU. The Lidar is positioned at the top, with the IMU located directly beneath it. The IMU provides 9-DOF inertial measurements including rigid body orientation, angular velocity, and acceleration.  These data are used in Lidar odometry optimization and Lidar points deskewing. The four wide-angle cameras are utilized to capture RGB features for 3D scene reconstruction. 

The four cameras are parameterized using fisheye models. We calibrate them by employing OpenCV~\cite{intcalibraion}, where the intrinsic and distortion parameters of each camera are computed based on a calibration checkerboard.
We then utilize the approach proposed in~\cite{extcalibraion} to perform extrinsic parameter calibration, which describes the relationship between the Lidar and the four cameras.
 

An FSYNC/FSIN (frame sync) signal is utilized for time synchronization among multiple camera sensors, operating at 10 Hz, which results in the same capture frequency per camera.
This sync signal consists of a pulse that goes high at the beginning of each frame capture to trigger all four camera shutters simultaneously.

The Ouster Lidar system works at 10 Hz, while the UM7 IMU provides data at a rate of 200 Hz. Unlike the hardware synchronization methods employed between cameras, the synchronization between the IMU and Lidar, as well as between the cameras and Lidar, is achieved solely through software, where timestamp data is utilized to align the outputs from the various sensors. Software synchronization is convenient and cost-effective, whereas the relatively noisy timestamps may result in less-than-optimal IMU pre-integration in SLAM and cause misalignment in point-cloud colorization.

\begin{figure*}[!h]
    \centering
    \includegraphics[width=1\linewidth]{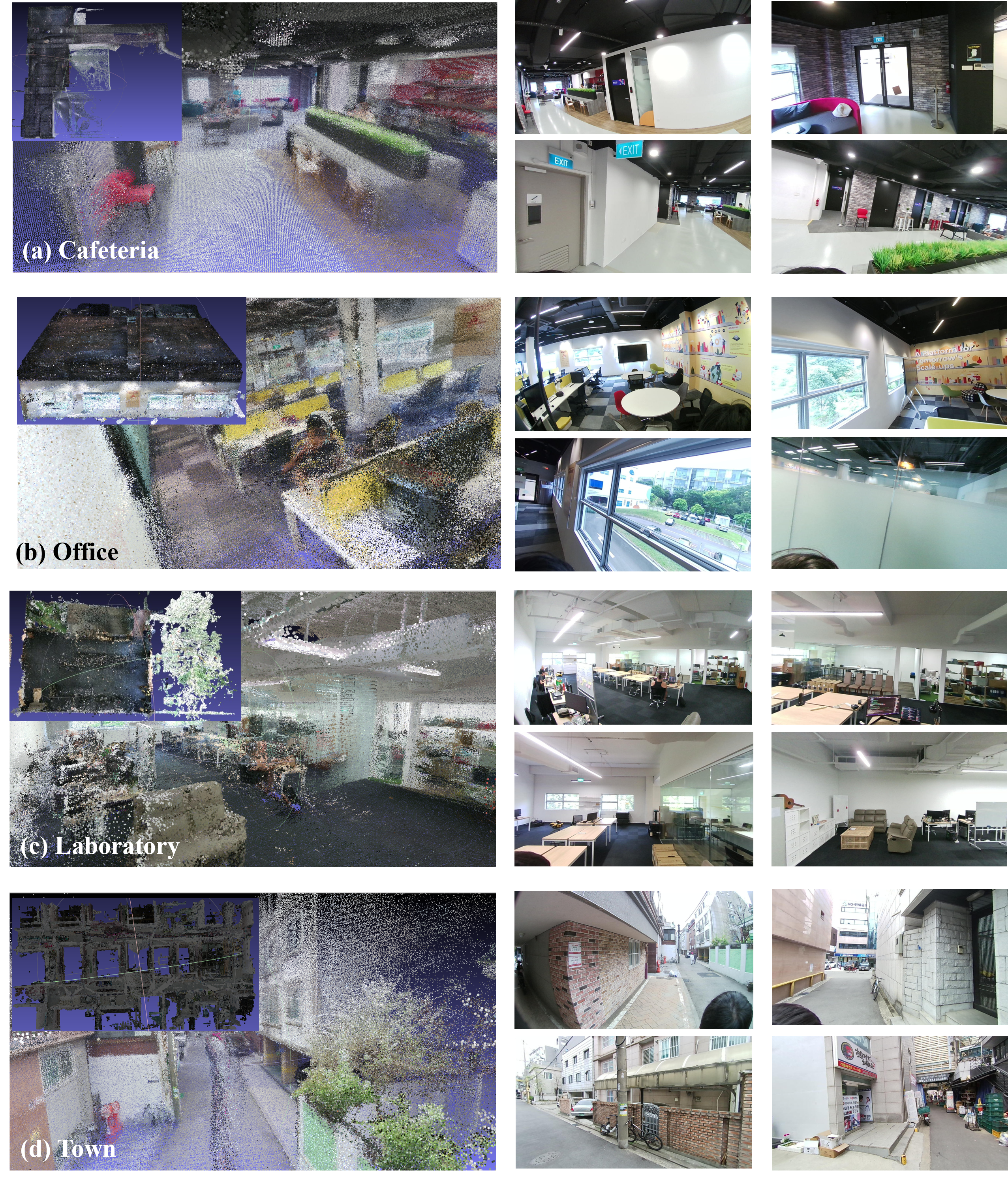}
    \caption{Qualitative examples of our proposed dataset. With our self-developed SLAM hardware system, we capture a new dataset comprising four scenes, including Cafeteria, Office, Laboratory and Town.}
    \label{fig:dataset}
\end{figure*}

With this SLAM setup, we present a new dataset which covers various environments, including three complex indoor scenes as well as a large-scale outdoor scene. We show a qualitative overview of this dataset in Fig.~\ref{fig:dataset} and present the key statistics in Table~\ref{tab:dataset_stat}.

    

\paragraph{Data pre-processing:}\label{subsec:data_preprocessing}
We first undistort the wide-angle images based on the estimated intrinsic and distortion parameters from camera calibration, which produces prospectively correct images with a large FoV of 97\textdegree. To reduce any influence of dynamic objects on our 3D reconstruction process, we employ a publicly available \texttt{Yolo v8} model \cite{reis2023real} that detects and spatially localizes passengers in these images. We exclude all pixels within their bounding boxes from further processing.

Our dataset offers dense point clouds for each scene, with a point number of 4-57 million points. As they often overpass the GPU memory limitation, we downsample the point cloud to a voxel size of 0.05 m in all experiments.

\section{Derivation of intrinsic refinement}\label{apd:intrinsic}
Most research employing 3DGS assumes the prior availability of accurate camera intrinsic parameters~\cite{matsuki2024gaussian, fan2024instantsplat, keetha2024splatam}. However, this assumption is difficult to fulfill, especially with SLAM devices that are equipped with multiple wide-angle cameras. Inaccurate intrinsic parameter estimates often lead to blurred reconstructed images, particularly at the image boundaries, as shown in Fig.~7 of the main paper. This issue is most severe in setups with multiple cameras, significantly degrading the quality of reconstruction. Despite its importance, this problem is frequently overlooked by the research community. We tackle this by enhancing the 3DGS rasterizer to refine imprecise camera intrinsic parameters during joint reconstruction optimization. This enhancement is achieved through an analytical solution, where the backward pass of the rasterization can be expressed as:
\begin{align*}
    \partialn{\mathcal{L}}{f_x} = \partialn{\mathcal{L}}{u} \times \partialn{u}{f_x}, &\quad     \partialn{\mathcal{L}}{f_y} =  \partialn{\mathcal{L}}{v} \times \partialn{v}{f_y};  \\
    \partialn{\mathcal{L}}{c_x} = \partialn{\mathcal{L}}{u} \times \partialn{u}{c_x}, &\quad  \partialn{\mathcal{L}}{c_y} =  \partialn{\mathcal{L}}{v} \times \partialn{v}{c_y}.
\end{align*}
Following the chain rule, the initial terms in each equation are the partial derivatives from the loss to the $uv$ variables, representing the screen coordinates of Gaussian ellipses. These derivatives are precomputed using the differentiable rasterizer. The subsequent terms are the derivatives of $uv$ with respect to the intrinsic parameters, which have analytical solutions expressed as:
\begin{align*}
    \partialn{u}{f_x} = \nicefrac{\vec{u}^{x}_{\rm{cam}}}{\vec{u}^{z}_{\rm{cam}}};  & \quad  \partialn{u}{c_x} = 1 \\
    \partialn{v}{f_y} = \nicefrac{\vec{u}^{y}_{\rm{cam}}} {\vec{u}^{z}_{\rm{cam}}};  &\quad 
    \partialn{v}{c_y} = 1 
\end{align*}
where $\vec{u}_{\rm{cam}}$ represents the Gaussian mean in camera space, with its components $\vec{u}^{x}_{\rm{cam}}$, $\vec{u}^{y}_{\rm{cam}}$ and $\vec{u}^{z}_{\rm{cam}}$ corresponding to the $x, y$ and $z$ dimensions, respectively.

\section{Exposure compensation module}\label{apd:exposure}

Our captures were taken in \textit{uncontrolled} settings, where significant variations in lighting conditions exist during the data acquisition. Training directly with these images can introduce the floaters and degrade the scene geometry \cite{darmon2024robust, lin2024vastgaussian}. To address this, we introduce an efficient exposure compensation module to handle issues related to illumination and exposure, drawing inspiration from \cite{Reiser2023SIGGRAPH} and \cite{wang2024bilateral}. We hypothesize that the variations in illumination are region-specific and affect the image's brightness in a gradual manner. Thus, our objective is simply to correct the illumination aspect of the images using a \textit{learnable} and \textit{low-frequency} offset. 

\begin{figure}
    \centering
    \includegraphics[width=0.99\linewidth]{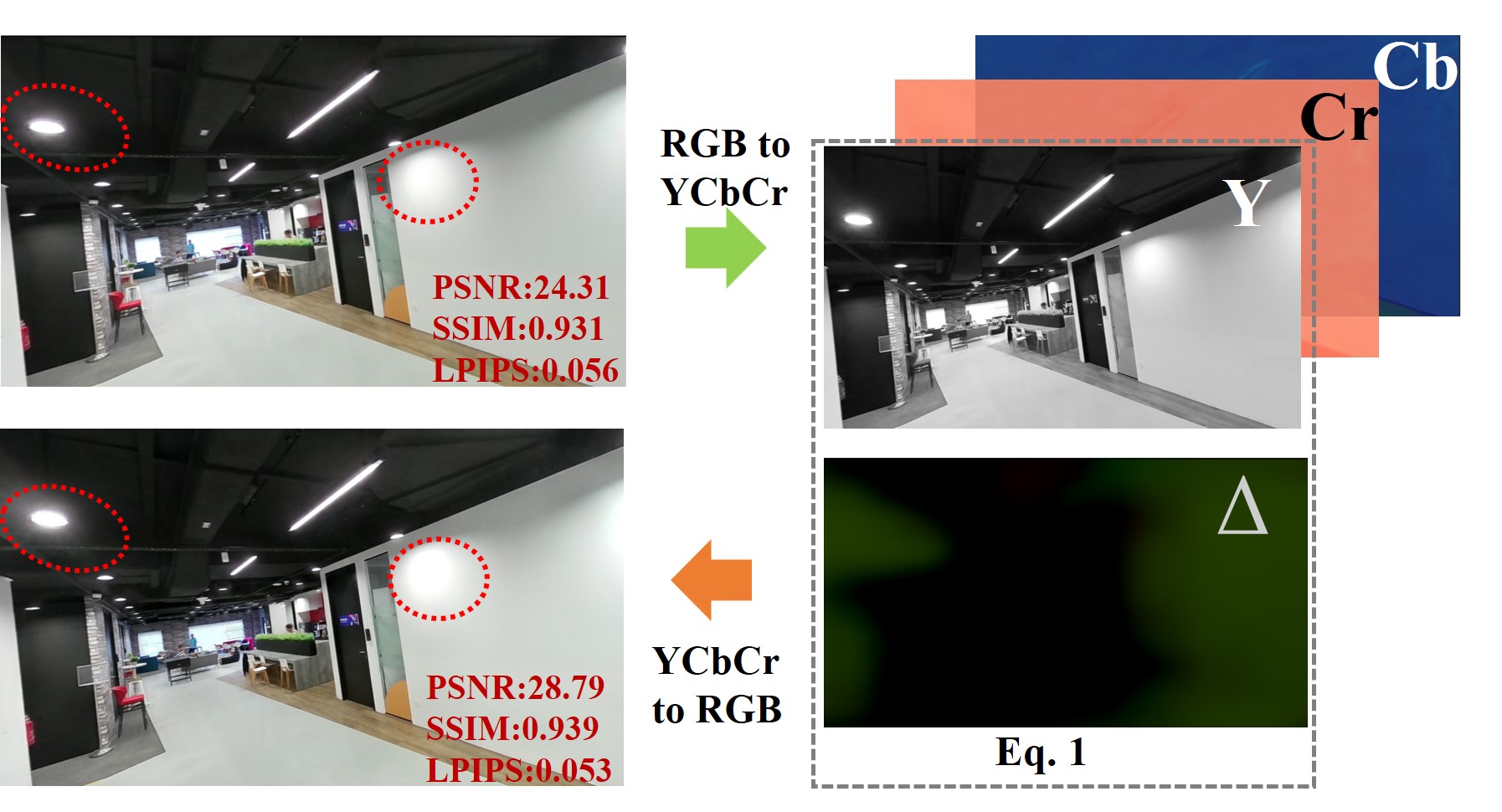}
    \caption{Illustration on our proposed exposure compensation module. In this approach, we project the image into YCbCr color space and only modify the channel representing illumination with a learnable low-frequency $\Delta$. We observe that $\Delta$ mainly highlights strong lighting regions.}
    \label{fig:exposure-pipeline}
\end{figure}

In particular, for an image $I\in \mathbb{R}^{3\times h\times w}$, we initially transform it from the RGB color space to the YCbCr color space \cite{noda2007colorization}, denoted as  $I_{\rm{YCbCr}}\in \mathbb{R}^{3\times h\times w}$. In this transformed space, the first dimension $I_{\rm{Y}}\in \mathbb{R}^{1\times h\times w}$ corresponds to the image luminance, representing brightness. The second and third dimensions, $I_{\rm{Cb}}$ and $I_{\rm{Cr}}$, capture the chrominance, thereby defining the color context of the image. 
Our learnable offset $\Delta^{2\times h\times w}$ is applied on the luminance dimension of the image as a small affine transformation. This compensates for region-specific inconsistency caused by either lighting condition or auto-exposure changes, as follows:
\begin{equation}
I_{\rm{Y}}^{'} = \Delta_{[0:1]} \times I_{\rm{Y}} + \Delta_{[1:2]}.
\end{equation}
We then obtain our resulting image
by projecting the $I_{\rm{YCbCr}}$ with modified $I_{\rm{Y}}^{'}$ back to the original RGB color space. 
The parameter $\Delta$ is defined per training image and is generated by a compact neural network implemented using \texttt{tinycudann}~\cite{mueller2022instant}. This network consists of a multi-resolution hash-encoding grid and a one-layer MLP.
We set \texttt{n_lelves} to 2 with a base resolution of $8\times8$, which ensures that $\Delta$ can only capture coarse spatial information. More importantly, we further smooth $\Delta$ with a low-pass Gaussian filter with a large kernel size of $51\times 51$ pixels. We illustrate our exposure compensation scheme in Fig.~\ref{fig:exposure-pipeline}. 

Since these offsets are not available for test images, we learn $\Delta$ per test image during the test-time optimization, together with the refinement of camera poses.

\section{Line intersection-based depth estimation}\label{subsec:line_intersec}
We compute the depth of two matched key-point pair by determining the intersection point of two lines defined by the camera origins and their view directions. 
We first consider two lines, $l_{1}$ and $l_2$, in 3D space, with origins $\vec{o}_1$ and $\vec{o}_2$, and directions $\vec{d_1}$ and $\vec{d_2}$, respectively.
Our objective is to find the points on lines $\vec{l}_1$ and $\vec{l}_2$, parameterized by the scalars, $t$ and $s$:
\begin{align}
    \vec{l}_1(t) &= \vec{o}_1 + t\cdot \vec{d}_1 \nonumber,\\
    \vec{l}_2(s) &= \vec{o}_2 + s\cdot \vec{d}_2, \nonumber   
\end{align}
so that the distance between these two points $||\vec{l}_2(s)-\vec{l}_1(
t)||^2$ are minimized (0 if the two lines intersect).

A necessary condition for this minimization is that the vector $\vec{l}_2(s)-\vec{l}_1(t)$ must be perpendicular to both $\vec{d}_1$ and $\vec{d}_2$, which can be expressed as:
\begin{align}
    \left(\vec{l}_2(s)-\vec{l}_1(t)\right)\cdot \vec{d}_1 &= (\vec{x}_{21} + s\cdot \vec{d}_2 - t\cdot \vec{d}_1) \cdot \vec{d}_1 =0, \nonumber \\
    \left(\vec{l}_2(s)-\vec{l}_1(t)\right)\cdot \vec{d}_2 &= (\vec{x}_{21} + s\cdot \vec{d}_2 - t\cdot \vec{d}_1) \cdot \vec{d}_2 =0, \nonumber
\end{align}

\noindent where $\vec{x}_{21} = \vec{o}_2-\vec{o}_1$ denotes the vector between the two origins.
These conditions can be derived by setting the first-order gradient of the distance function to zero.
By applying these two conditions, one can obtain the analytic solution, resulting in the following expressions:
\begin{equation}
    t = \frac{||\vec{d}_2||^2 \cdot \vec{x}_{21}\cdot \vec{d}_1 - \vec{x}_{21}\cdot \vec{d}_2\cdot (\vec{d}_1 \cdot \vec{d}_2)   }{||\vec{d}_1\cdot \vec{d}_2||^2 - ||\vec{d}_1||^2 ||\vec{d}_1||^2} ,
\label{eq:ts}
\end{equation}
\begin{equation}
    s = -\frac{||\vec{d}_1||^2 \cdot \vec{x}_{21}\cdot \vec{d}_2 - \vec{x}_{21}\cdot \vec{d}_1\cdot (\vec{d}_1 \cdot \vec{d}_2)   }{||\vec{d}_1\cdot \vec{d}_2||^2 - ||\vec{d}_1||^2 ||\vec{d}_1||^2}.
\label{eq:ts2}
\end{equation}
In our setting, for each pair of matched points, the origins $\vec{o}_1$ and $\vec{o}_2$ are defined as the camera centers. The directions $\vec{d}_1$ and $\vec{d}_2$ represents the vectors from these camera centers towards their respective image planes, determined by the $uv$ coordinates, image dimensions, and intrinsic parameters. Note that the camera origins and directions are expressed in the world coordinates. Given $t$ and $s$, we can compute the depth of these two matched pixels by applying the viewing matrix and extracting the z-axes element. We disregard matched pairs where $t$ or $s$ is zero or negative, as the line intersection must be in front of both cameras. Additionally, we ignore pairs with very small angles (less than 2\textdegree) between $\vec{d}_1$ and $\vec{d}_2$, as this makes Eqs.~\ref{eq:ts} and~\ref{eq:ts2} unstable due to very small denominators.

\section{Extended implementation details and discussion}\label{apd:extended}

In this section, we provide implementation details of our proposed constrained-optimization based method, as well as the comparison approaches. 


All experiments were conducted on a machine equipped with an Intel-14900K CPU and an NVIDIA 4090 GPU. Our framework is based on the open-source differentiable rasterizer~\cite{kerbl20233d, ye2024gsplatopensourcelibrarygaussian}, with modifications to accommodate non-centric images, enable differentiable depth rendering, and ensure differentiability in both extrinsic and intrinsic parameters. To facilitate optimization and avoid sub-optimal solutions, we employed a cosine learning rate decay strategy with restarts. Specifically, we increased the learning rate and performed the decay three times during the optimization process, starting at the $1^{\rm{st}}$, $\nicefrac{\rm{\texttt{max_iter}}}{6}$, and $\nicefrac{\rm{\texttt{max_iter}}}{2}$ iterations. Considering that the point clouds roughly capture the scene geometry, we disabled the pruning operation during optimization for all experimental variants, while enabling Gaussian point densification starting $67\%$ of its training.

\begin{figure}[t]
    \centering
    \includegraphics[width=1\linewidth]{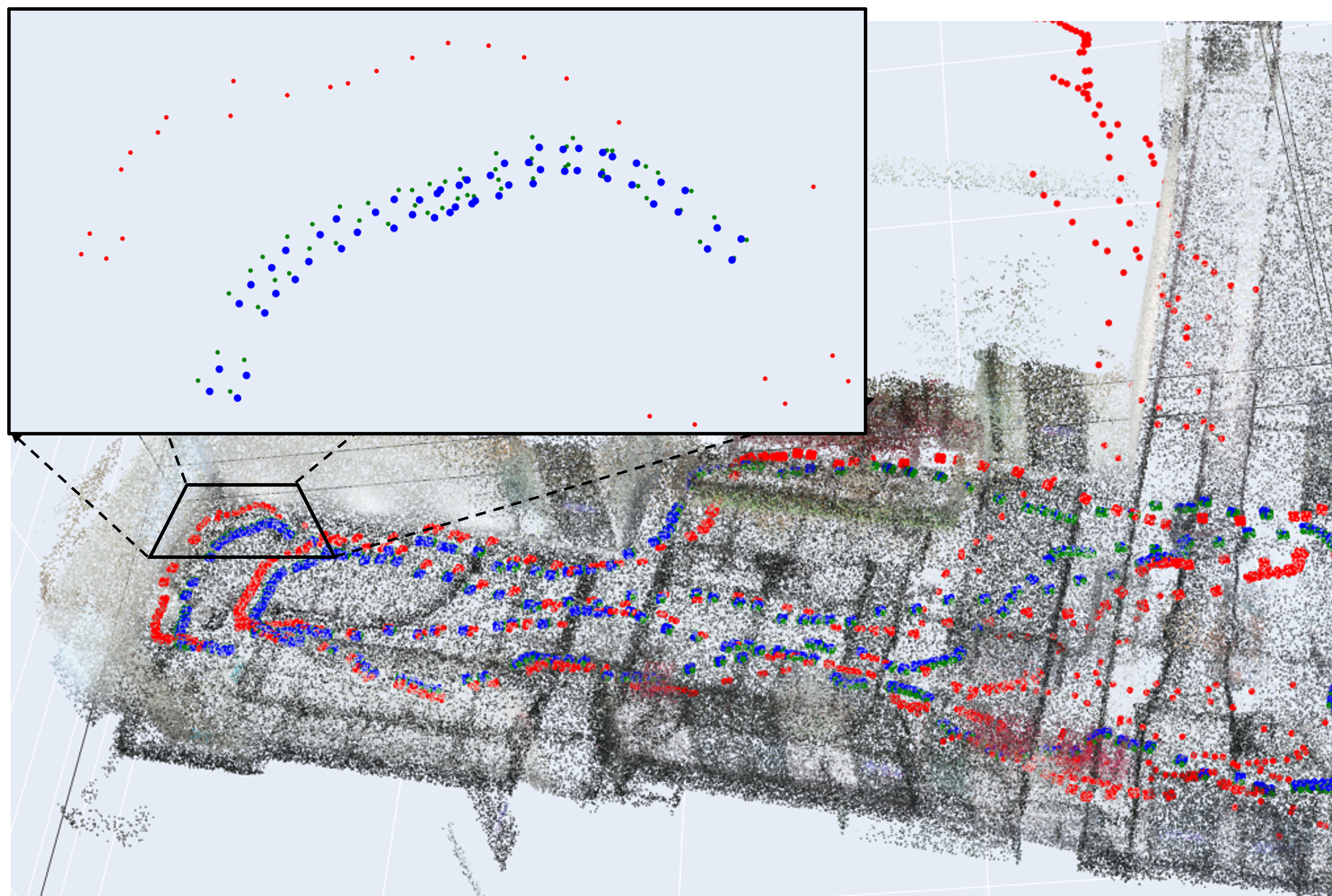}
    \caption{Camera pose visualization for the Cafeteria scene. {\color{red} Red} and {\color{ForestGreen}green} points represent the trajectories estimated by 3DGS-COLMAP and 3DGS-COLMAP$^\triangle$, respectively. 3DGS-COLMAP fails to capture the geometry structure, while our method, shown in {\color{blue}blue}, converges very similarly to 3D-COLMAP$^\triangle$ but with better visual quality (as shown in Table 1 in the main paper). }
    \label{fig:colmap-failure}
\end{figure}

\begin{figure}[b]
    \centering
    \includegraphics[width=0.95\linewidth]{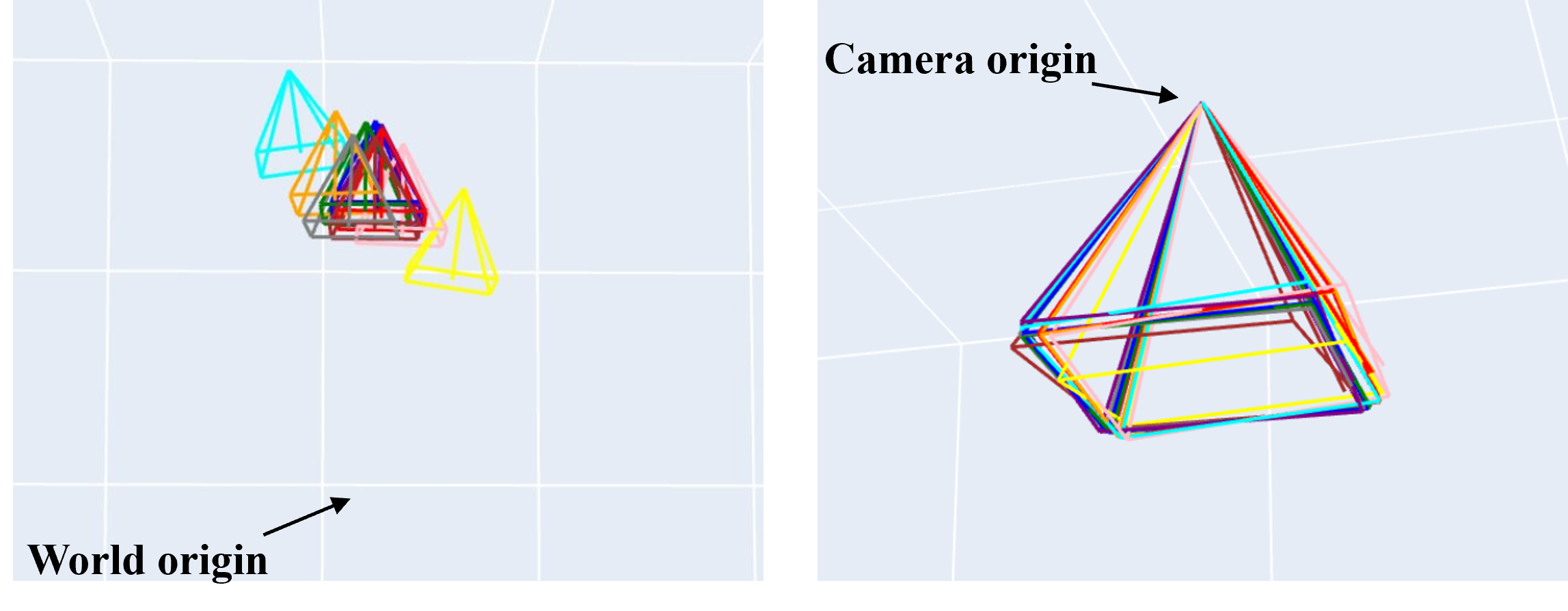}
    \caption{
    Qualitative comparison for two camera pose optimization approaches. Different colors represent camera poses at various time during optimization. \textbf{Left:} camera poses are optimized by rotating around the world origin (Eq. 6 in main paper). \textbf{Right:} camera poses are rotated around the initial camera origin (Eq. 7 in main paper). Our proposed approach on the right demonstrates better optimization robustness.
    }
    \label{fig:camera_error}
\end{figure}
In the following, we provide the implementation details on comparing methods:
\begin{itemize}
\item \textbf{3DGS-COLMAP}: We enhanced this widely-used baseline by associating the camera information with RGB images. This was achieved by modifying the \texttt{database} file of generated by COLMAP software, with the intrinsic estimations as a prior.

\item \textbf{3DGS-COLMAP$^{\triangle}$}: The next method takes the initial camera poses as additional priors and perform rig-based bundle adjustment. This is achieved using COLMAP's \texttt{point_triangulator} and \texttt{rig_bundle_adjuster} interfaces.

 \item  \textbf{CF-3DGS}~\cite{Fu_2024_CVPR}: This approach incrementally estimates the camera poses based solely on visual images, which utilizes two distinct Gaussian models: a local model and a global model. The local Gaussian model calculates the relative pose differences between successive images, while the global Gaussian model aims to model the entire scene and refine the camera poses derived from the local model. Since this approach is designed for a monocular camera configuration and requires video-like input, we provide our images on a per-camera basis to ensure compatibility. Unexpectedly, this method failed to capture the geometry after processing approximately 10 images, resulting in significantly poor rendering. This issue is primarily due to our key frames having a moderate covisibility threshold. Additionally, the frames exhibit a repetitive block pattern and feature plain surfaces in many scenes, which impede this visual-based method from accurately estimating the camera poses.
 
 
\item \textbf{MonoGS} \cite{matsuki2024gaussian}: 
Similar to the previous approach, this technique incrementally reconstructs the scene while simultaneously estimating camera positions by optimizing for photometric loss and depth inconsistency loss. In our experiments, we found that this baseline faces a similar challenge as CF-3DGS, specifically, a difficulty in accurately capturing the true geometry from a diverse and uncontrolled set of images. Consequently, it produces entirely empty images after processing about 15 images across all tested scenes. We interrupt and restart the training when it fails completely, continuing this process until the method can provide a test score on our designated set of test images
 
\item \textbf{InstantSplat}~\cite{fan2024instantsplat}: 
This approach uses 3D foundation models to generate a dense and noisy point cloud, which is then optimized along with the camera extrinsics. Originally designed for sparse-view synthesis, we found it challenging to handle more than 30 images due to GPU memory limitations. To adapt this method to our context, we selected a sequence of 30 images, consisting of 29 consecutive training images and one test image strategically placed in the middle. The individual test score is computed on the sub-model, which requires one minute of pre-processing and 50 seconds of training time. We report the test score based on the average of multiple sub-models.

\item \textbf{LetsGo}~\cite{cui2024letsgo}: 
Similar to ours, this approach proposed to integrate high-quality point cloud and camera poses with enhanced 3DGS technology. We follow their open-sourced implementation\footnote{\url{https://github.com/zhaofuq/LOD-3DGS}}, default training parameters, and test it on different sequences of GarageWorld and Waymo datasets.
\item \textbf{StreetGS}~\cite{yan2024street}: 
The last multi-modality method aims to reconstruct dynamic driving scenes with dynamic and static Lidar point clouds and high quality camera poses. Similar to our 3DGS-COLMAP baselines, this baseline first refines the camera poses using COLMAP and then optimize each camera pose independently during the reconstruction. We follow the default setting in their open-sourced implementation\footnote{\url{https://github.com/zju3dv/street_gaussians}} to test this method on both Waymo and GarageWorld datasets, except that we only reconstruct the static scene while ignoring the moving objects.   

\end{itemize}

We present in Fig.~\ref{fig:colmap-failure} the pose estimation results from 3DGS-COLMAP, 3DGS-COLMAP$^\triangle$, and our method. As illustrated, 3DGS-COLMAP fails in this scene due to repeated block patterns and plain surfaces. We also show in Fig.~\ref{fig:camera_error} the qualitative examples for two different camera pose refinement approaches. We observe that using the Eq.~7 in the main paper results in a more stable optimization trajectory.

\section{Extended experimental setup and results on \texttt{GarageWorld} dataset}
\label{sec:letsgo}
\begin{figure}
    \centering
    \includegraphics[width=1\linewidth]{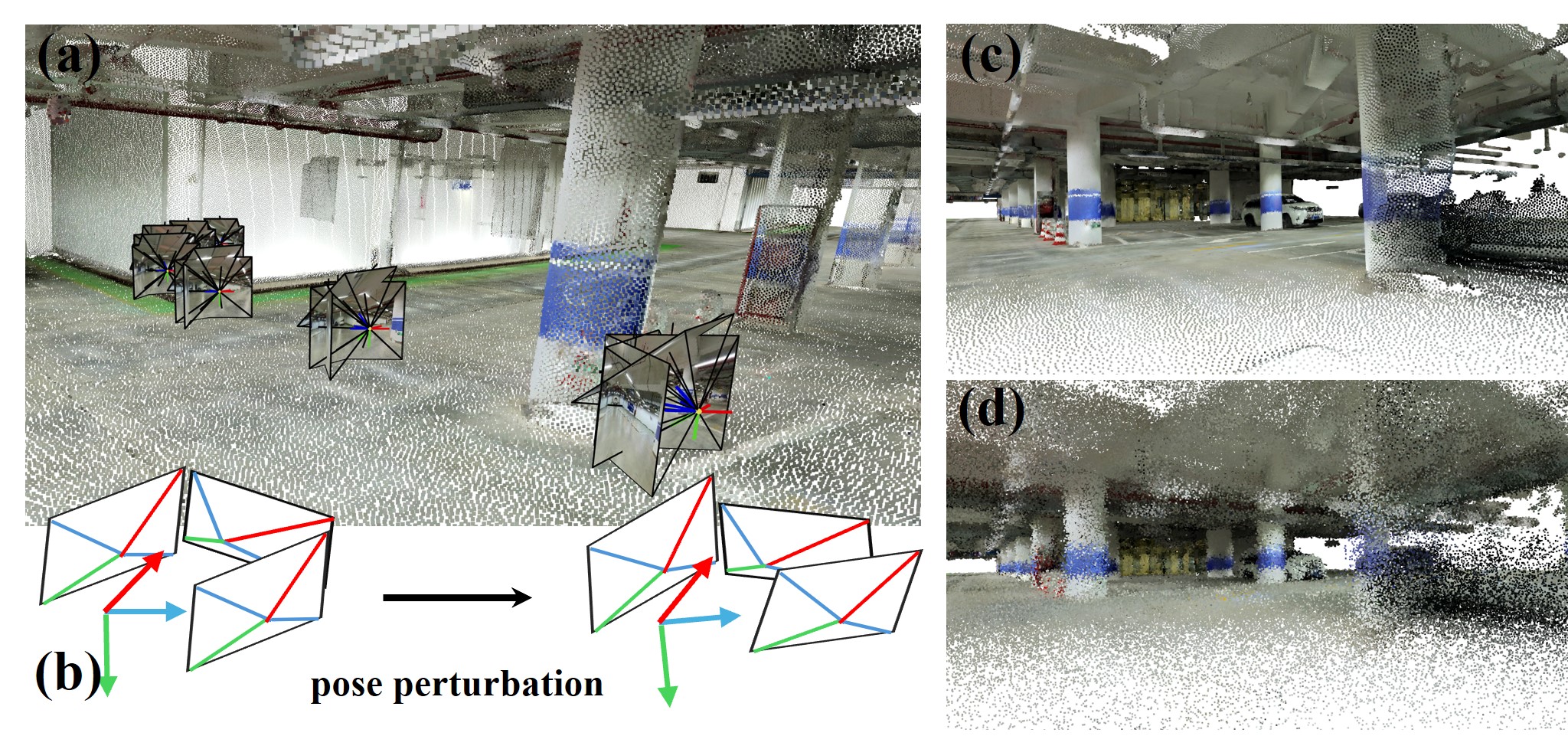}
    \caption{Illustration of the \texttt{Garage World} dataset with four undistorted cameras oriented in various directions. (b) We perturbed the camera poses using pose decomposition. (c) and (d) show the point cloud both before and after the introduction of perturbations.}
    \label{fig:dataset-perturbation}
\end{figure}
\begin{figure*}
    \centering
    \includegraphics[width=1\linewidth]{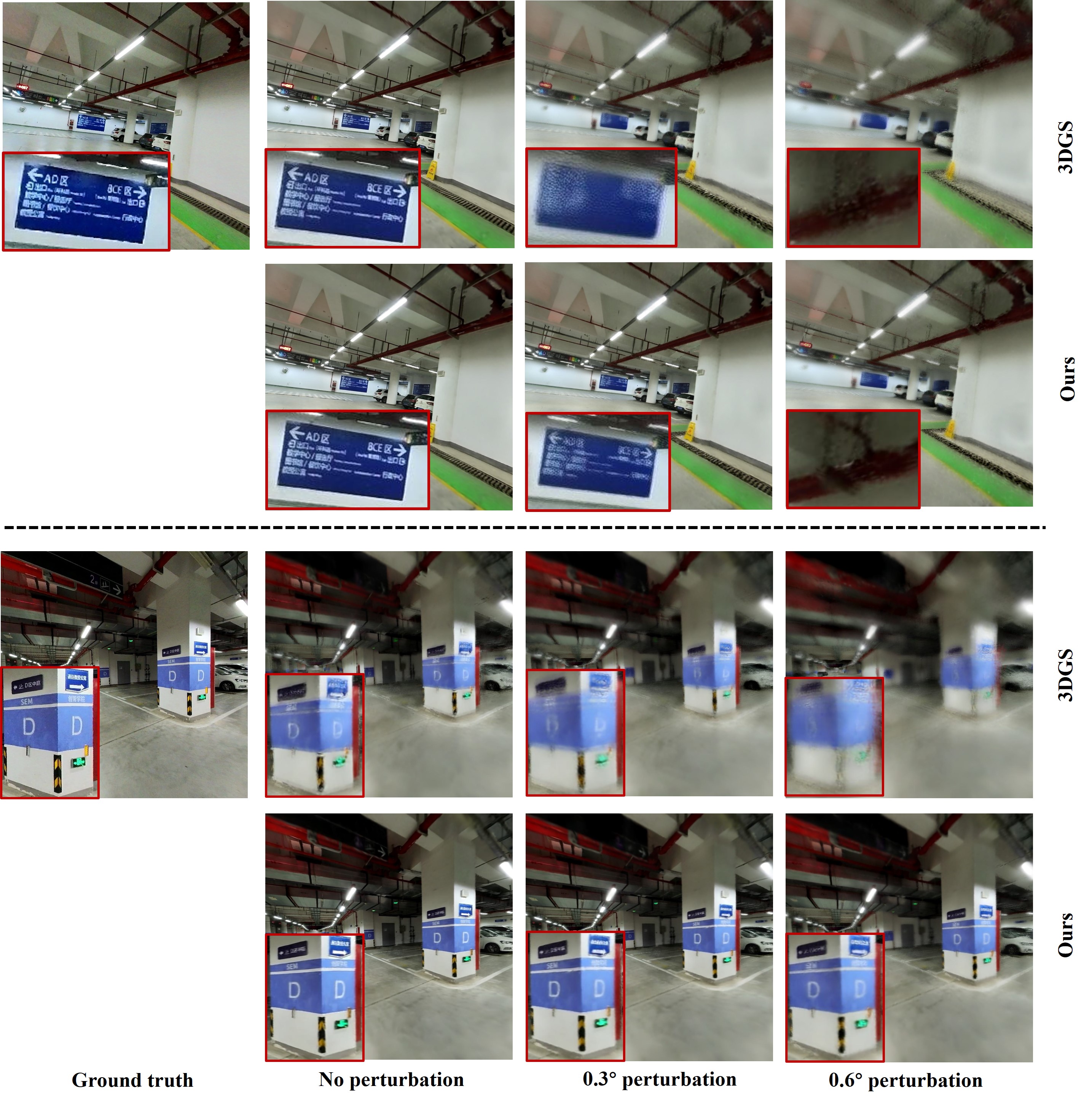}
    \caption{Qualitative comparison of our constrained optimization approach with the 3DGS baseline. The top and bottom respectively show clean and perturbed scenes for groups 0 and 6 at different levels. We show that our method enhances visual quality in the presence of camera pose errors and maintains better quality even without noise injection.}
    \label{fig:garage-world-visual-result}
\end{figure*}

\begin{table}
\centering
\caption{Quantitative comparisons on the (perturbed) Garage World dataset. We show that our proposed method can consistently improve the performance despite large perturbations.}
\label{tab:garage-world}
\begin{adjustbox}{max width=1\linewidth}
\setlength{\tabcolsep}{2pt}
\renewcommand{\arraystretch}{1}
\begin{tabular}{c|c|ccc|ccc}
    \toprule
    \multirow{2}{*}{\begin{minipage}{0.3in} \textbf{Noise Level} \end{minipage}}
    & \multirow{2}{*}{\textbf{Method}} 
    & \multicolumn{3}{c}{\textbf{Group 0}} & \multicolumn{3}{|c}{\textbf{Group 6}}
			\\
    \cmidrule(l{0pt}r{0pt}){3-8} 
    & & \bfseries PSNR $\uparrow$ & \bfseries SSIM $\uparrow$ & \bfseries LPIPS $\downarrow$ & \bfseries PSNR $\uparrow$ & \bfseries SSIM $\uparrow$ & \bfseries LPIPS $\downarrow$ \\
    \midrule
    \multirow{2}{*}{\textbf{-}} & \textbf{3DGS} & 25.43 & 0.8215 & 0.2721 &21.23 & 0.7002 & 0.4640 \\
    & \textbf{Ours} & \cellcolor{red!25} 26.06 & \cellcolor{red!25} 0.8325 & \cellcolor{red!25} 0.2606 & \cellcolor{red!25}23.76& \cellcolor{red!25} 0.7779 &\cellcolor{red!25}0.3537 \\

    \midrule
    \multirow{2}{*}{\textbf{0.3\textdegree}} & \textbf{3DGS} &  23.17 & 0.7595 & 0.4033 & 21.00 & 0.6979 & 0.5085 \\
    & \textbf{Ours} & \cellcolor{red!25}25.12 & \cellcolor{red!25}0.8060 & \cellcolor{red!25}0.3110 & \cellcolor{red!25}23.06 & \cellcolor{red!25}0.7515 & \cellcolor{red!25}0.4004\\

    \midrule
    \multirow{2}{*}{\textbf{0.6\textdegree}} & \textbf{3DGS} & 22.07 & 0.7388 & 0.4645 & 20.58 & 0.6874 & 0.5359 \\
    & \textbf{Ours} & \cellcolor{red!25}23.09 & \cellcolor{red!25}0.7594 & \cellcolor{red!25}0.3995 & \cellcolor{red!25}21.94 & \cellcolor{red!25}0.7160 & \cellcolor{red!25}0.4611\\

    \bottomrule
		
\end{tabular}
\end{adjustbox}
\end{table}
We are particularly interested in \texttt{GarageWorld}~\cite{cui2024letsgo} dataset due to its high relevance to our work. We conducted extensive experiments on this dataset to validate the robustness of our proposed method.
Unlike our collected dataset, this dataset provides \textit{highly accurate} camera poses and \textit{very clean} point cloud but with only one fisheye camera. Fortunately, four pinhole images are undistorted from the same wide-angle image with fixed view directions: Front, Left, Right, and Up, as shown in Fig.~\ref{fig:dataset-perturbation} (a).
We therefore consider this dataset as an image collection from multiple-camera setup and decompose the camera poses into device-center and camera-to-device transformations. We further test our method against 3DGS baseline~\cite{kerbl20233d} on two sequences, {Group 0} and {Group 6}, randomly drawn from the campus scene. This extended experimental results are shown in both Table~\ref{tab:garage-world} and Fig.~\ref{fig:garage-world-visual-result}.

The first two rows of Table~\ref{tab:garage-world} show the experimental results under the ideal conditions. Due to the high-quality camera poses and the clean point cloud, the reconstruction performance for 3DGS reaches a PSNR score of 25.43 and 21.23 dB for both scenes. Notably, our proposed method consistently outperforms the baseline across both scenes and all visual metrics. 
The rendered test images exhibit clearer edges and more detailed context, which can be attributed to our method's ability to mitigate even subtle intrinsic and extrinsic errors encountered during time-intensive acquisitions with complex hardware.


Our next series of experiments aim to demonstrate the robust capabilities of our proposed method using a dataset with varying levels of perturbation. To achieve this, we introduce Gaussian noise to both the device-center and camera-to-device poses, as well as to the point cloud, creating synthetic datasets with degradations. This process is illustrated in Fig.~\ref{fig:dataset-perturbation} (b) and (d).

The third and fourth rows of Table~\ref{tab:garage-world} present experimental results under conditions of mild degradation. Both camera-to-device and device transformations were adjusted with a random Gaussian noise limited to 0.3\textdegree in their orientations. Additionally, random Gaussian noise confined to 0.01 m was added to the initial point cloud. This noise negatively affects the 3DGS baseline performance, whereas our proposed method shows quality improvements by 1.95 dB, 4.65\%, and 9.23\% across the three visual criteria. In the second scene, there is an enhancement of 2.06 dB, 5.36\%, and 10.8\%. 

The final two rows in Table~\ref{tab:garage-world} represent the performance of both methods under greater perturbations, with orientations disturbed up to 0.6\textdegree. Our method enhances reconstruction performance in both evaluated scenes, particularly for the LPIPS metric, and maintains credible rendering quality despite the challenging conditions.

\section{Qualitative comparison on Waymo dataset}
\label{sec:waymo}
\begin{figure*}
    \centering
    \includegraphics[width=1\linewidth]{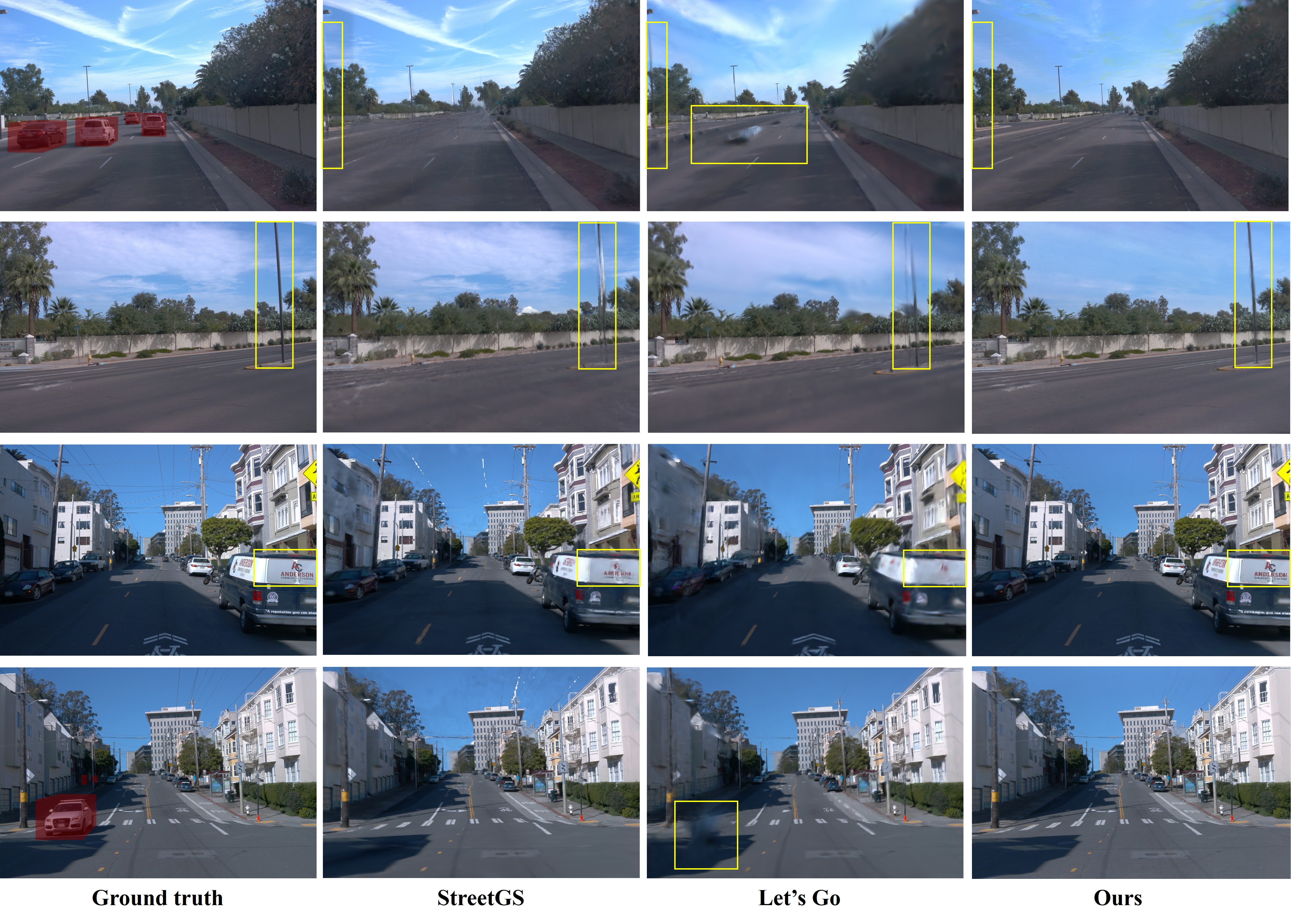}
    \caption{Qualitative comparison of our constrained optimization approach with multimodal methods. We overlay the dynamic object mask on the ground truth images to highlight the static regions on which our metrics are computed. We show that our proposed method offers better scene geometry and rendering details compared with state-of-the-art approaches.}
    \label{fig:waymo-visual-result}
\end{figure*}

As shown in Fig.~\ref{fig:waymo-visual-result}, we present qualitative comparisons of our proposed method against state-of-the-art multimodal 3DGS approaches which integrate cameras, Lidars, and inertial sensors. We show that our method can better reconstruct scene geometries, as evidenced by straight rendered streetlights, and achieves a higher level of detail in the final rendering. These improvements demonstrate the effectiveness of our approach in capturing fine-grained structural and textural information, leading to a more realistic and visually consistent representation of the scene.

\end{document}